\documentclass[10pt,twocolumn,letterpaper]{article}

\usepackage[pagenumbers]{iccv} 

\definecolor{iccvblue}{rgb}{0.21,0.49,0.74}
\usepackage[pagebackref,breaklinks,colorlinks,allcolors=iccvblue]{hyperref}
\usepackage{algorithm}
\usepackage{algorithmic}
\usepackage{bm}
\usepackage{amsthm}
\newtheorem{remark}{Remark}

\def\paperID{14106} 
\def\confName{ICCV}
\def\confYear{2025}

\title{Addressing Representation Collapse in Vector Quantized Models with \\One Linear Layer}

\author{Yongxin Zhu$^{1,2}$, Bocheng Li$^{1,2}$, Yifei Xin$^{3}$, Zhihua Xia$^{4}$, Linli Xu\thanks{Corresponding author.}$^{~~1,2}$\\
$^{1}$University of Science and Technology of China \\
$^{2}$State Key Laboratory of Cognitive Intelligence, $^{3}$Peking University, $^{4}$Jinan University\\
\texttt{zyx2016@mail.ustc.edu.cn,bcli@mail.ustc.edu.cn}\\
\texttt{xinyifei@stu.pku.edu.cn,xiazhihua@jnu.edu.cn,linlixu@ustc.edu.cn}
}

\begin{document}
\maketitle
\begin{abstract}
Vector Quantization (VQ) is essential for discretizing continuous representations in unsupervised learning but suffers from representation collapse, causing low codebook utilization and limiting scalability. Existing solutions often rely on complex optimizations or reduce latent dimensionality, which compromises model capacity and fails to fully solve the problem. We identify the root cause as disjoint codebook optimization, where only a few code vectors are updated via gradient descent. To fix this, we propose \textbf{Sim}ple\textbf{VQ}, which reparameterizes code vectors through a learnable linear transformation layer over a latent basis, optimizing the \textit{entire linear space} rather than nearest \textit{individual code vectors}. Although the multiplication of two linear matrices is equivalent to applying a single linear layer, this simple approach effectively prevents collapse. Extensive experiments on image and audio tasks demonstrate that SimVQ improves codebook usage, is easy to implement, and generalizes well across modalities and architectures. The code is available at \url{https://github.com/youngsheen/SimVQ}.
\end{abstract}

\section{Introduction}

In recent years, vector quantization (VQ) \cite{NIPS2017_7a98af17,NEURIPS2019_5f8e2fa1} has emerged as a foundational technique in unsupervised representation learning \cite{NEURIPS2020_92d1e1eb,pmlr-v235-bruce24a} and latent generative models \cite{Rombach_2022_CVPR,yu2022vectorquantized,yu2022scaling,10158503,wang2023neural,zhu-etal-2024-generative}. By converting continuous representations into discrete codes, VQ models can effectively identify the inherent structure of data and enable various discrete modeling methods on continuous data, from high-quality image generation \cite{Esser_2021_CVPR} to audio synthesis \cite{efossez2023high}. The recent success of Large Language Models (LLMs)~\cite{achiam2023gpt} has highlighted the effectiveness of next-token prediction as a powerful and versatile training objective. Consequently, VQ models are taken as the direct method to transform data from various modalities \cite{zhang-etal-2023-speechgpt,sun2024autoregressive,team2024chameleon} or scientific domains \cite{gao2024foldtoken} to discrete sequences for next token prediction training. However, attempts to integrate VQ models as multimodal tokenizers to leverage the scaling laws of LLMs face significant challenges because of the difficulty of expanding the codebook. For example, the Chameleon model \cite{team2024chameleon} constrains its codebook size to $8k$, which is significantly trailing behind the vocabulary size of LLMs (e.g., LLaMA3's vocabulary size is $128k$ \cite{dubey2024llama}).

\begin{figure*}[t]
    \centering
    \includegraphics[width=2.0\columnwidth]{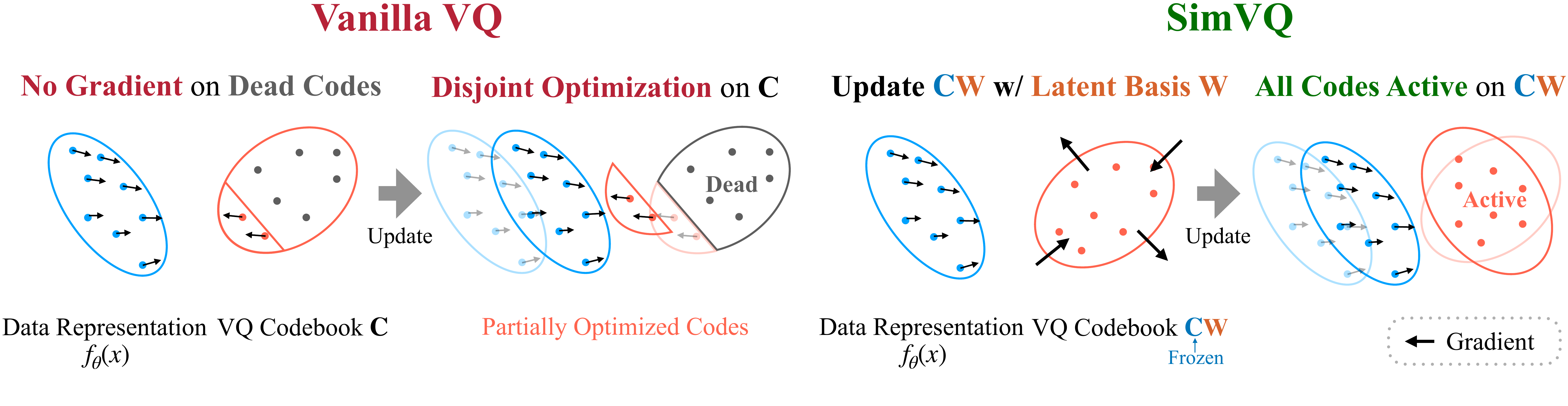}
    \caption{\textbf{Comparison of Vanilla VQ and SimVQ.} (a): (left) Disjoint optimization in Vanilla VQ. Only the nearest codes are updated, resulting in a high percentage of ``dead'' codes that are not updated. (b): (right) Joint optimization in SimVQ. The entire codebook is updated with a latent basis, ensuring all codes remain active.}
    \label{fig:intro}
\end{figure*}

There is a broad agreement that increasing vocabulary size can consistently improve the performance of LLMs \cite{tao2024scaling}. However, recent studies \cite{Zhu2024ScalingTC} indicate that traditional VQ models often fail to utilize the additional parameters introduced by codebook expansion, leaving most codes inactive during training. The contradiction between codebook expansion and low codebook utilization in VQ models is known as the representation collapse problem \cite{roy2018theory}, where increasing the codebook size fails to improve the performance. To address these discrepancies, we conduct a theoretical analysis of the optimization procedure of VQ models and identify that the disjoint optimization of the codebook is the root cause of representation collapse. As illustrated in Fig. \ref{fig:intro}(a), the core mechanism of VQ models involves a nearest-neighbor replacement strategy, where the encoder’s output features are replaced by the nearest vector in the codebook to serve as input to the decoder. The indices of the nearest vector are taken as the discrete representation of the data. This nearest-selection operator results in only a subset of codes being updated through gradient descent, while the remaining codes remain unchanged.

Some methods mitigate representation collapse by hand-designing complex optimization strategies, such as stochastic quantization \cite{pmlr-v162-takida22a}, distribution penalty \cite{VQWasserstein,Xiao2023SCVAESC}, and codebook reset \cite{zheng2023online} through sophisticated training strategies. 
Recently, some approaches \cite{yu2022vectorquantized,mentzer2024finite,yu2024language} propose to reduce the dimension of the latent space to a very small scale (e.g., 8 v.s. 128) to alleviate the curse of dimensionality, thereby improving the overlap between the encoder’s features and the codebook. However, while these methods enhance codebook utilization, they do so at the cost of model capacity, leading to worse performance compared to vanilla VQ models when the codebook size is small or representation collapse is not severe. Another approach, VQGAN-LC \cite{Zhu2024ScalingTC}, initializes the codebook with features extracted from the pre-trained CLIP model \cite{pmlr-v139-radford21a} to create a well-structured latent space that better matches the distribution of the encoder output. Nevertheless, the latent space defined by an external pre-trained model limits the model's ability to generalize to diverse datasets and reaches a performance plateau as the codebook size increases. These limitations highlight the need for a more effective method to improve codebook utilization without compromising model capacity or relying on external models.

We critically assess prevalent methodologies and reveal that optimizing the latent space rather than individual code vectors is key to preventing representation collapse. Building on this insight, we introduce a simple yet effective method, termed SimVQ, to directly update the latent space spanned by the codebook by linear transforming the code vectors via a learnable latent basis. Specifically, the vectors in the codebook are reparameterized as a linear combination of the basis in the learnable linear layer $\bm{W}$:
\begin{equation}
    \bm{C} \in \mathbb{R}^{K\times d} \Rightarrow \bm{CW} ~\text{with}~ \bm{W} \in \mathbb{R}^{d\times d},
\end{equation}
where $K$ denotes the codebook size and $d$ represents the dimension of latent space. This reparameterization with linear transformation disentangles the optimization of the codebook into two components: the coefficient matrix $\bm{C}$ and the basis of linear space $\bm{W}$ respectively. As illustrated in Fig. \ref{fig:intro}(b), by optimizing the basis matrix $\bm{W}$, the latent space spanned by $\bm{CW}$ is rotated and stretched to match encoder's output feature. The entire codebook is updated jointly to prevent the representation collapse problem. The simplicity of the proposed method makes it highly portable and easily adaptable for improving VQ-based models across a wide range of domains, requiring only \textit{one linear layer}.

In summary, our contributions to vector quantized models are as follows:
\begin{itemize}
    \item We theoretically analyze the representation collapse problem in VQ models and reveal that optimizing the latent space spanned by the codebook, rather than individual code vectors, is crucial to addressing this issue. 
    \item We propose a novel method, SimVQ, which reparameterizes the codebook vectors in VQ models via a linear transformation with a learnable latent basis. This simple yet effective approach is highly adaptable and easy to implement, making it broadly applicable across various machine learning contexts.
    \item We conduct an extensive evaluation of SimVQ across diverse modalities, including image and audio with different model architectures. The results show that SimVQ not only effectively addresses the representation collapse problem by achieving near-complete codebook utilization regardless of the codebook size, but also establishes new state-of-the-art performance. Furthermore, when scaling up the codebook size, SimVQ consistently delivers improved results.
\end{itemize}

\section{Related Work}
VQ-VAE \cite{NIPS2017_7a98af17} is the pioneering work to encode data into discrete representations. Building on these developments, VQGAN \cite{Esser_2021_CVPR} combines VQ-VAE with adversarial networks to improve the perceptual quality of generated samples and establish a fundamental quantization protocol in latent generative models \cite{Rombach_2022_CVPR,yu2022scaling,team2024chameleon}. 
Many traditional VQ works focus on training a better discrete representation rather than codebook utilization to improve reconstruction performance. For example, RVQ \cite{Lee2022AutoregressiveIG} and MoVQ \cite{Zheng2022HighQualityPI} enhance the reconstruction details with multichannel quantization. 
However, these methods suffer from a critical issue of representation collapse, as they struggle to scale the codebook size beyond 10k entries, limiting their scalability. In response to this challenge, several approaches have been proposed recently. 
DALLE \cite{pmlr-v139-ramesh21a} employs the gumbel-softmax trick \cite{DBLP:conf/iclr/JangGP17} and stochastic sampling strategies to activate most codes during training. However, during inference, only a small subset of codes is utilized for quantization \cite{Zhang_2023_CVPR}. 
Some methods design complex optimization strategies to improve codebook utilization, such as stochastic quantization \cite{pmlr-v162-takida22a}, distribution penalty \cite{VQWasserstein,Xiao2023SCVAESC} and codebook reset \cite{zheng2023online}.
\citet{pmlr-v202-huh23a} proposes rescaling the vectors in the codebook during training to match the distributions in the latent space. VQGAN-FC \cite{yu2022vectorquantized} introduces a method to map latent vectors into a lower-dimensional space followed by $l_2$ normalization to alleviate representation collapse. FSQ \cite{mentzer2024finite} extends this idea by projecting representations into a reduced-dimensional space, where they are quantized into a small set of fixed values. LFQ \cite{yu2024language}, a variant of FSQ, uses binary values for quantized representations, thereby simplifying the encoding process. While these methods improve the codebook utilization, they do so at the cost of model capacity by significantly reducing the dimensionality of latent space (often to as low as 8), leading to worse performance compared to vanilla VQ models when the codebook size is small and representation collapse is not severe. Additionally, VQGAN-LC \cite{Zhu2024ScalingTC} proposes to initialize the codebook using features extracted from the pre-trained CLIP model to avoid representation collapse. However, the reliance on the pre-trained model limits the VQ model's ability to generalize to diverse datasets and results in a performance plateau as the codebook size increases. In contrast, our method, SimVQ, effectively addresses the representation collapse problem with a simple linear layer, without sacrificing model capacity or relying on external pre-trained models.

\section{Representation Collapse in VQ Models}

\subsection{Preliminaries}

A vector quantized model is typically a reconstructive encoder-decoder architecture that includes a vector quantization layer to convert continuous representations into discrete codes. For simplicity, we represent an image with a single random variable $x$. Formally, the encoder $f_{\theta}$ maps the input image into a latent space, producing a continuous representation $z_{e}=f_{\theta}(x) \in \mathbb{R}^d$. This representation is then quantized using a learnable codebook $\bm{C}=[q_1,\ldots,q_K]\in \mathbb{R}^{K\times d}$, where $q_i$ is a codebook vector. We define $\delta_{k}\in \{0,1\}^{1\times K}$ as a characteristic (one-hot) vector where only the $k$-th element is $1$, such that $q_k = \delta_{k}\bm{C}\in \mathbb{R}^{1\times d}$. The quantization layer selects the nearest codebook vector $q_k$ by minimizing the Euclidean distance between $z_e$ and the codebook entries \cite{NIPS2017_7a98af17}:
\begin{align}
    k = \arg\min_{j} \|z_{e}-q_j\|^2_2 = \arg\min_{j} \|z_{e}-\delta_{j}\bm{C}\|^2_2.
\end{align}
The selected vector $q_k$ is then passed to the decoder $g_{\phi}$ to reconstruct the input image. 

To enable gradient propagation through the non-differentiable characteristic vector $\delta_{k}$, the straight-through estimator (STE) \cite{bengio2013estimating} is applied. During the backward pass, the gradient of $z_q=\delta_k \bm{C}$ is copied to $z_e$ as follows,
\begin{align}
    z_q = \text{sg}(\delta_{k}\bm{C} - z_e) + z_e,\quad \Rightarrow \frac{\partial z_q}{\partial z_e}=1
\end{align}
where \texttt{sg} is the stop gradient operator, ensuring the gradient for $\delta_k \bm{C}$ is discarded during the backward pass.

The learning objective is the combination of a reconstruction loss and commitment loss that ensures that the encoder commits to an embedding and the encoder's output does not drift:
\begin{equation}
    \mathcal{L} = \log p(x|z_q) + \|\text{sg}(\delta_{k}\bm{C})-z_e\|^2_2 + \beta \|\delta_{k}\bm{C} - \text{sg}(z_e)\|^2_2,
\end{equation}
where $\log p(x|z_q)$ is typically the mean squared error (MSE) loss $\|x-g_{\phi}(z_q)\|^2_2$ for image and audio data.

\subsection{Disjoint Optimization of Codebook}

In VQ models, only the nearest code is selected and updated via gradient descent. Ideally, all codebook entries should be updated and utilized for decoding. However, experimental evidence shows that only a small fraction of the codebook gets updated and utilized, leading to what is known as the representation collapse problem \cite{roy2018theory}. To investigate the root cause of this issue, we provide a theoretical analysis of the optimization dynamics in VQ models.

Due to the use of the straight-through estimator (STE) for gradient propagation, the codebook $\bm{C}$ can only be updated through the gradient of the commitment loss, which is defined as:
\begin{equation}
    \mathcal{L}_{commit}(\bm{C})=\|z_e - \delta_k\bm{C}\|_2^2.
\end{equation}
The codebook $\bm{C}$ is updated according to the following equation, where $\eta$ is the learning rate:
\begin{align}
    &\bm{C}^{(t+1)} = \bm{C}^{(t)} + \eta \mathbb{E}_{z_e} \left[\frac{\partial \mathcal{L}_{commit}(\bm{C}^{(t)})}{\partial \bm{C}^{(t)}}\right] \\
    &= \bm{C}^{(t)} - \eta \mathbb{E}_{z_e} \left[\delta_k^T\delta_k\bm{C}^{(t)}\right] + \eta \mathbb{E}_{z_e} \left[\delta_{k}^{T}z_e\right]
\end{align}
where $\delta_k^T\delta_k$ is the Kronecker delta matrix, defined as:
\begin{equation}
    (\delta_k^T \delta_k)_{ij} =
\begin{cases}
1 & \text{if } i = j = k, \\
0 & \text{otherwise}.
\end{cases}
\end{equation}
All vectors in $\bm{C}$ will be updated and utilized if and only if the expectation $\mathbb{E}_{z_e}\left[\delta_k^T\delta_k \right]$ converges to the identity matrix. Unlike variational autoencoders (VAEs) \cite{Kingma2013AutoEncodingVB}, which enforce a Gaussian distribution on the latent space via a KL-divergence penalty, VQ models optimize $z_e$ towards the selected codebook vectors $\mathbb{E}_{z_e} \left[\delta_k^T\delta_k\bm{C}\right]$. At the same time, the selected codebook vectors are optimized towards the distribution of $z_e$, resulting in the same selected subset of vectors moving closer to $z_e$, somewhat akin to a cocoon effect. However, this disjoint optimization of the codebook leads to part of the codebook, specifically $(\bm{I} - \mathbb{E}_{z_e}\left[\delta_k^T\delta_k \right])\bm{C}$, remaining un-updated and underutilized once the optimization process begins. This phenomenon occurs because the optimization focuses only on a subset of codebook vectors, leaving other vectors stagnant.

This analysis reveals the fundamental cause of representation collapse in VQ models: the disjoint optimization process that updates only a subset of codebook vectors. This insight forms the basis for our proposed solution, SimVQ, which aims to address this issue by optimizing the entire latent space spanned by the codebook, rather than individual code vectors.

\section{Addressing Collapse with Latent Linear Transformation}

\subsection{Reparameterize Codes with Latent Basis}

Let $\bm{W}=\{\bm{w}_1,\ldots ,\bm{w}_n\}$ be a basis of a linear space. Any vector $\bm{v}$ in the space can be uniquely expressed as a linear combination of the basis vectors with coefficients $c_1,\ldots,c_n \in \mathbb{R}$:
\begin{equation}
    \bm{v} = c_1\bm{w}_1+\cdots+c_n\bm{w}_n = \bm{c}\bm{W}.
\end{equation}
Given the equivalence between $\bm{v}$ and $\bm{c}\bm{W}$ in the linear space, we can reparameterize each vector in the codebook of VQ models with a new basis matrix $\bm{W}\in \mathbb{R}^{d\times d}$. Specifically, the codebook $\bm{C}=\{\bm{c}_1,\ldots,\bm{c}_K\}$ can be reparameterized as:
\begin{equation}
    \{\bm{\hat{c}}_1\bm{W},\ldots,\bm{\hat{c}}_N\bm{W}\}=\bm{\hat{C}}\bm{W}\in \mathbb{R}^{K\times d}.
\end{equation}
This reparameterization introduces two components: the basis matrix $\bm{W}$ and the coefficient matrix $\bm{\hat{C}}$. In the following, we will discuss the optimization of both the basis matrix $\bm{W}$ and the coefficient matrix $\bm{\hat{C}}$. For simplicity, we slightly abuse $\bm{C}$ and $\bm{\hat{C}}$ below.

\subsection{Asymmetric Optimization Dynamics}

While it is commonly accepted that multiplying two linear matrices is equivalent to a single linear layer, we argue that the disjoint optimization problem of the codebook in VQ models can be addressed by linear transformation. In vanilla VQ models, only the codebook $\bm{C}$ is responsible for minimizing commitment loss, leading to the disjoint optimization problem where only the selected codes will be updated. 

In contrast, when the codebook is reparameterized as $\bm{C}\bm{W}$, both the basis $\bm{W}$ and the coefficient matrix $\bm{C}$ contribute to minimizing the commitment loss. The gradients $\frac{\partial \mathcal{L}}{\partial \bm{W}}$ and $\frac{\partial \mathcal{L}}{\partial \bm{C}}$ can simultaneously reduce the loss. As a result, the optimization of the reparameterized codebook can be divided into three scenarios:
\begin{itemize}
\item Updating $\bm{C}$ with $\bm{W}$ frozen: Only the \textbf{selected} codes adapt to the latent distribution of $z_e$, as depicted on Fig. \ref{fig:intro}(a). The vanilla VQ is a special case with $\bm{W}=\bm{I}$.
\item Updating $\bm{W}$ with $\bm{C}$ frozen: The \textbf{entire} codebook $\bm{C}\bm{W}$ adjusts to the latent distribution of $z_e$. The basis matrix $\bm{W}$ rotates and stretches the space as shown in Fig. \ref{fig:intro}(b).
\item Updating both $\bm{C}$ and $\bm{W}$: The selected subset of codes moves towards $z_e$ while the space spanned by $\bm{W}$ undergoes simultaneous rotation and stretching.
\end{itemize}

To highlight the difference in optimization between $\bm{C}$ and $\bm{CW}$, we conduct a toy experiment in a two-dimensional setting and visualize the optimization process in Fig. \ref{fig:optim1} and Fig. \ref{fig:optim2}. 

\subsubsection{Toy Examples}

\begin{figure}[t]
    \centering
    \begin{minipage}[b]{0.49\columnwidth}
        \centering
        \includegraphics[width=\columnwidth]{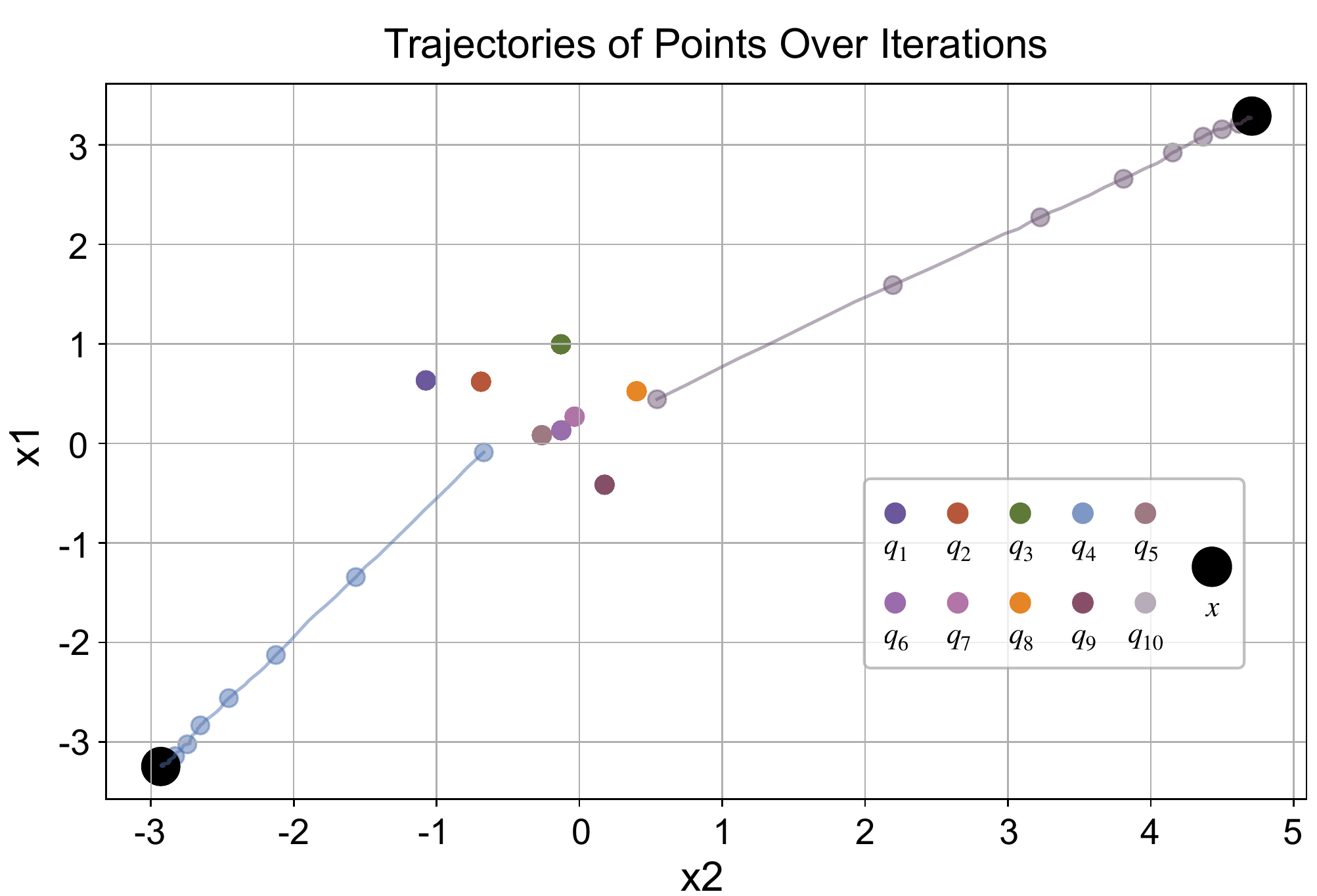}
    \end{minipage}
    \begin{minipage}[b]{0.49\columnwidth}
        \centering
        \includegraphics[width=\columnwidth]{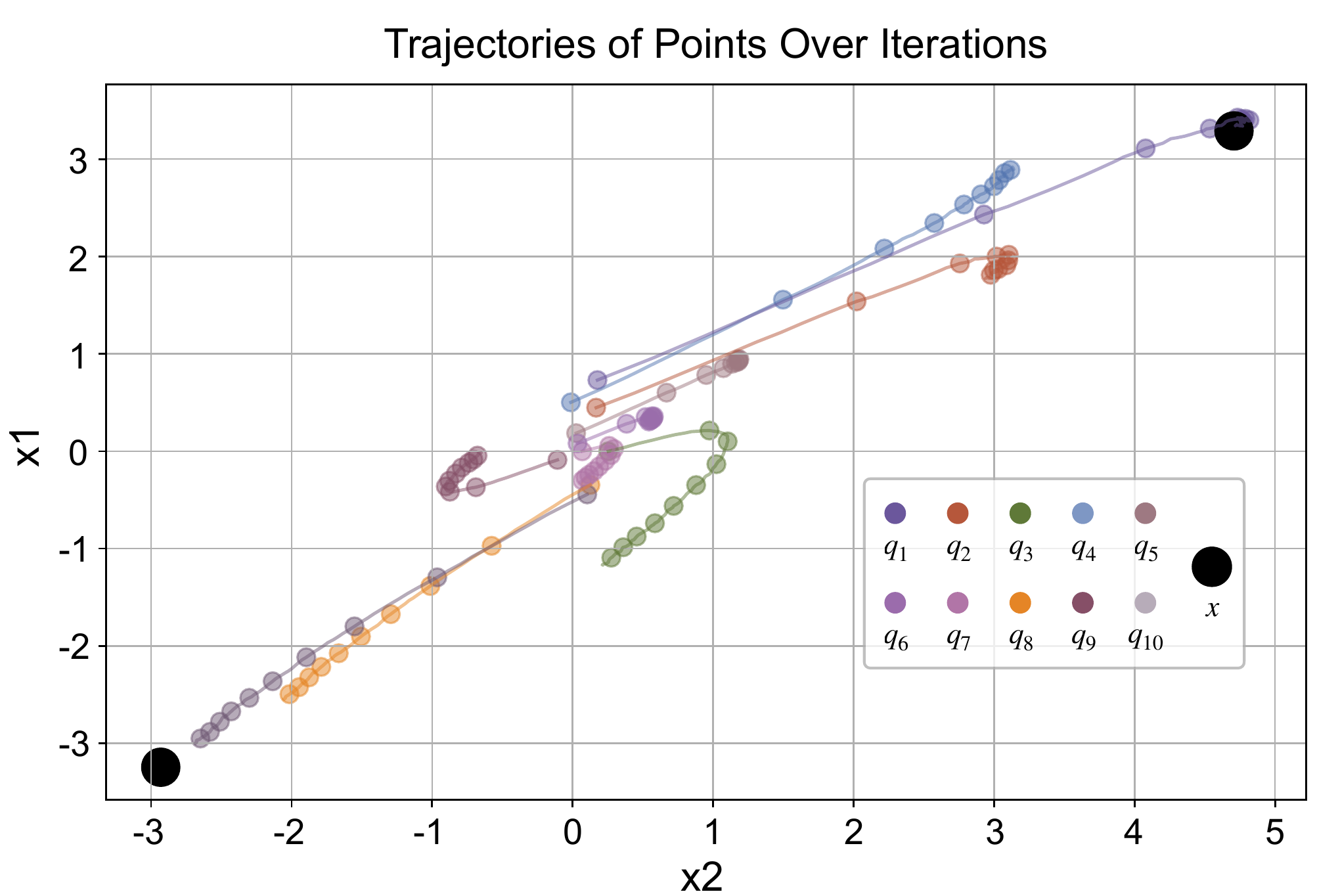}
    \end{minipage}
    \caption{(a): (left) The optimization trajectory of the objective $\|\bm{x}-\bm{q}\|^2_2$, which is the same as vanilla VQ. Only a small fraction of points are updated while others remain inactive. (b): (right) The optimization trajectory of the objective $\|\bm{x}-\bm{q}\bm{w}\|^2_2$ with $\bm{q}$ frozen, which is the same as SimVQ. All the points are updated towards targets $x$.}
    \label{fig:optim1}
\end{figure}

\begin{figure}[t]
    \centering
    \begin{minipage}[b]{0.47\columnwidth}
        \centering
        \includegraphics[width=\columnwidth]{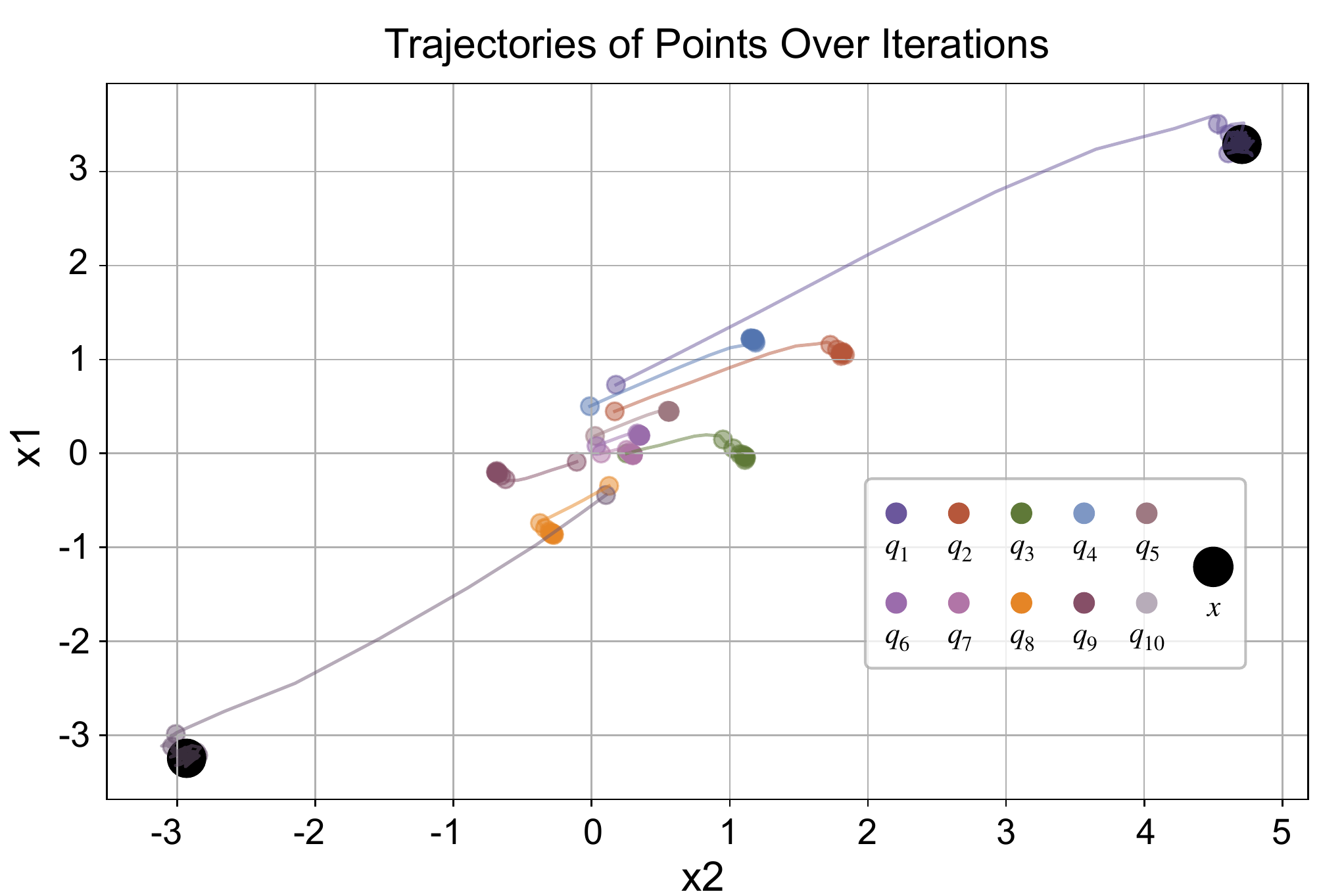}
    \end{minipage}
    \begin{minipage}[b]{0.51\columnwidth}
        \centering
        \includegraphics[width=\columnwidth]{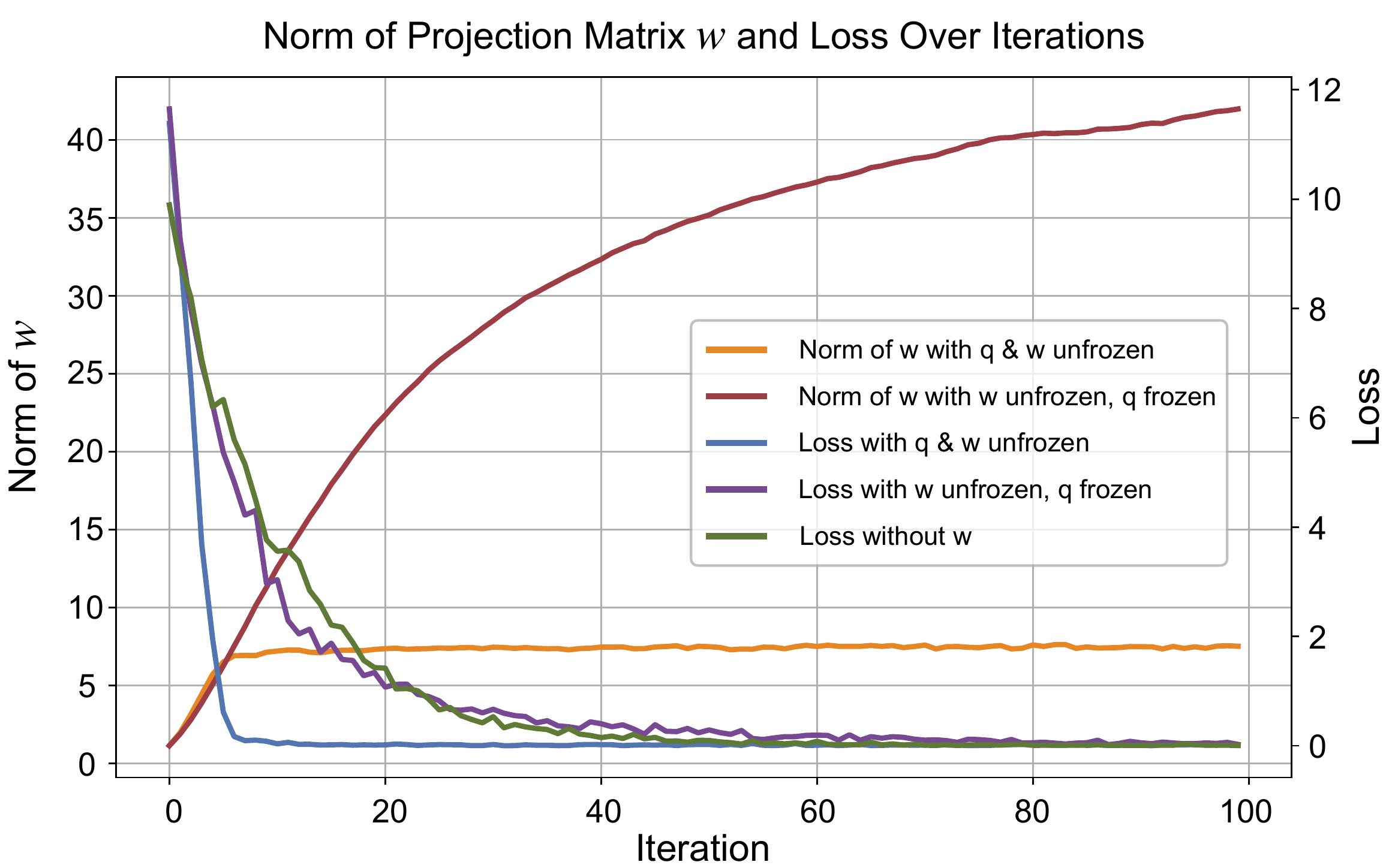}
    \end{minipage}
    \caption{(a): (left) The optimization trajectory of the optimization objective: $\|\bm{x}-\bm{q}\bm{w}\|^2_2$ with both $\bm{q}$ and $\bm{w}$ unfrozen. (b): (right) The Frobenius norm of the projection matrix $\bm{w}$ and loss curves. The loss quickly converges to 0 with $\bm{w}$ almost unchanged.}
    \label{fig:optim2}
\end{figure}

\begin{algorithm}[t]
   \caption{Training Procedure 
   for SimVQ}
   \label{alg:code}
\begin{algorithmic}
\STATE \textbf{Input:} Encoder $f_{\theta}$, Decoder $g_{\phi}$, Codebook $\bm{C}\in \mathbb{R}^{K\times d}$, Linear projector matrix $\bm{W}_{\psi}$, commitment weight $\beta$. 
\STATE \textbf{Output:} Model parameters $\theta, \phi, \psi$ and Codebook $\bm{C}$.
\STATE Initialize Codebook with Gaussian distribution and \textbf{freeze} the parameter of Codebook $\bm{C}$;
\REPEAT
    \STATE Draw $x\sim p_{data}(\bm{x})$;
    \STATE $z_e=f_{\theta}(x)$;
    \STATE \textcolor{blue}{/* Replace $q_j$ in vanilla VQ with proposed $q_j\bm{W}_{\psi}$.}
    \STATE Nearest code search: $k = {\arg\min}_{j} \|z_e - \textcolor{blue}{q_j\bm{W}_{\psi}}\|^2_2$, where $q_j\in \bm{C}$; 
    \STATE Straight Through Estimation: $z_q=\text{sg}(\textcolor{blue}{q_k\bm{W}_{\psi}} - z_e)+z_e$;
    \STATE $\hat{x}=g_{\phi}(z_q)$;
    \STATE Minimize $\mathcal{L}(\theta,\phi,\psi)$=$\text{MSE}(x,\hat{x})+\beta \|z_e-\text{sg}(\textcolor{blue}{q_k\bm{W}_{\psi}})\|^2_2+\|\text{sg}(z_e)-\textcolor{blue}{q_k\bm{W}_{\psi}}\|^2_2$;
\UNTIL{converged}
\end{algorithmic}
\end{algorithm}

We randomly sample two target points $\bm{x}$ from Gaussian distribution as follows:
\begin{equation}
    \bm{x}_1 \sim \mathcal{N}(
    \left(
    \begin{matrix}
    2\\
    2
    \end{matrix}
    \right), 
    \left(
    \begin{matrix}
    1 & 0\\
    0 & 1
    \end{matrix}
    \right)
    ),\quad \bm{x}_2 \sim \mathcal{N}(
    \left(
    \begin{matrix}
    -2\\
    -2
    \end{matrix}
    \right), 
    \left(
    \begin{matrix}
    1 & 0\\
    0 & 1
    \end{matrix}
    \right)
    ).
\end{equation}
Then we initialize $10$ learnable vectors $\bm{q}$ from a Gaussian distribution:
\begin{equation}
    \{\bm{q}_i\}_{i=1}^{10} \sim \mathcal{N}(
    \left(
    \begin{matrix}
    0\\
    0
    \end{matrix}
    \right), 
    \left(
    \begin{matrix}
    1 & 0\\
    0 & 1
    \end{matrix}
    \right)
    ),
\end{equation}
During training with gradient descent, we introduce perturbation noise $\mathcal{N}(0,0.01)$ to the targets. In Fig. \ref{fig:optim1}(a), the optimization objective is similar to vanilla VQ: $\|\bm{x}-\bm{q}\|^2_2$. Only the nearest points $\bm{q}_4$ and $\bm{q}_{10}$ are updated. In contrast, in Fig. \ref{fig:optim1}(b), the optimization objective $\|\bm{x}-\bm{q}\bm{w}\|^2_2$ is similar to SimVQ with the points reparameterized by a learnable latent basis $\bm{w}$ and $\bm{q}$ frozen, resulting in the entire codebook $\{\bm{q}\}_{i=1}^{10}$ 
being \textit{jointly} updated.

\begin{remark}
The simultaneous optimization of the latent basis $\bm{w}$ and the coefficient matrix $\bm{q}$ \textbf{may} lead to the collapse. 
\end{remark}

We provide an example in Fig. \ref{fig:optim2}(a) where the optimization objective is $\|\bm{x}-\bm{q}\bm{w}\|^2_2$ with $\bm{q}$ unfrozen this time. In the training process, only the nearest point $\bm{q}_1$ and point $\bm{q}_{10}$ move towards the target point, while other points remain almost unchanged. We also visualize the loss curve in Fig. \ref{fig:optim2}(b). The optimization objective with both $\bm{q}$ and $\bm{w}$ unfrozen converges quickly, where the norm of basis $\bm{w}$ is much smaller than the objective with $\bm{q}$ frozen. This indicates that the disjoint optimization of the codebook persists: $\bm{q}$ can directly commit to the loss and dominate the optimization process, with $\bm{w}$ being ignored, leading to the collapse quickly.

\subsection{Joint Optimization of the Codebook}

We propose SimVQ by simply using a learnable basis $\bm{W}\in \mathbb{R}^{d\times d}$ to reparameterize the codebook such that the codebook 
is transformed into $\bm{C}\bm{W}$. The pseudo-code for this approach is provided in Algorithm \ref{alg:code}. During training, we optimize only the latent basis matrix $\bm{W}$, while keeping the coefficient matrix $\bm{C}$ frozen. The commitment loss for SimVQ is defined as:
\begin{equation}
    \mathcal{L}_{commit}(z_e,q_k)=\|z_e-\delta_k\bm{C}\bm{W}\|^2_2.
\end{equation}
The vanilla VQ model is a special case of SimVQ, where the linear basis matrix $\bm{W}$ is fixed as the identity matrix $\bm{I}$. The update for $\bm{W}$ with learning rate $\eta$ is:
\begin{align}
    &\bm{W}^{(t+1)} = \bm{W}^{(t)} - \eta \frac{\partial \mathcal{L}_{commit}(z_e,\bm{q}_k)}{\partial \bm{W}^{(t)}}\\ &= (\bm{I}-\eta \mathbb{E}_{z_e} \left[ \bm{C}^T\delta_k^T\delta_k\bm{C}\right])\bm{W}^{(t)} + \eta \mathbb{E}_{z_e} \left[ \bm{C}^T\delta_k^{T}z_e\right].
\end{align}
The term $\mathbb{E} \left[\bm{C}^T \delta_k^T \delta_k \bm{C}\right]$ represents the expectation of the quadratic form, and simplifies to $\mathbb{E}[\bm{q}_k^T \bm{q}_k]$. Since the codes are randomly sampled from a Gaussian distribution, we have:
\begin{equation}
    \mathbb{E}\left[\bm{q}_k^T\bm{q}_k\right]=\bm{I}, \text{where}~\bm{q}\sim \mathcal{N}(0,1),
\end{equation}
which ensures that all elements of $\bm{W}$ are updated. As training progresses, the latent basis $\bm{W}$ converges to: \begin{equation}
    \lim_{t\rightarrow \infty}\bm{W}^{(t)} = \mathbb{E}_{z_e} \left[\bm{q}_k^T z_e\right]
\end{equation}
Thus, in the limit:
\begin{equation}
    \lim_{t\rightarrow \infty} \bm{q}_k \bm{W}^{(t)}=\mathbb{E}\left[\bm{q}_k\bm{q}_k^T\bm{e}\right]=\mathbb{E}\left[\bm{e}\right]
\end{equation}
At convergence, the product $\bm{q}_k \bm{W}$ equals the nearest feature.

\begin{table*}[t]
\centering
\resizebox{1.85\columnwidth}{!}{
\begin{tabular}{lccccccc}
\toprule
 Method & Latent dim & Codebook size & Util$\uparrow$ & rFID$\downarrow$ & LPIPS$\downarrow$ & PSNR$\uparrow$ & SSIM$\uparrow$ \\
 \midrule
VQGAN \cite{Esser_2021_CVPR} & 128 & 65,536 & 1.4\% & 3.74 & 0.17 & 22.20 & 70.6 \\
VQGAN-EMA \cite{NEURIPS2019_5f8e2fa1} & 128 & 65,536 & 4.5\% & 3.23 & 0.15 & 22.89 & 72.3 \\
\midrule
VQGAN-FC \cite{yu2022vectorquantized}& 128 & 65,536 & 1.4\% & 5.33 & 0.18 & 21.45 & 68.8 \\
VQGAN-FC \cite{yu2022vectorquantized}& 8 & 65,536 & 100.0\% & 2.63 & 0.13 & 23.79 & 77.5 \\
FSQ$^\dagger$ \cite{mentzer2024finite} & 6 & 64,000 & 100.0\% & 2.80 & 0.13 & 23.63 & 75.8 \\
LFQ \cite{yu2024language} & 16 & 65,536 & 100.0\% & 2.88 & 0.13 & 23.60 & 77.2 \\
VQGAN-LC-CLIP$^+$ \cite{Zhu2024ScalingTC} & 768 & 65,536 & 100.0\% & 2.40 & 0.13 & 23.98 & 77.3 \\
\midrule
SimVQ (ours) & 128 & 65,536 & 100.0\% & \textbf{2.24} & \textbf{0.12} & \textbf{24.15} & \textbf{78.4} \\
SimVQ (ours) & 128 & 262,144 & 100.0\% & \textbf{1.99} & \textbf{0.11} & \textbf{24.68} & \textbf{80.3} \\
\bottomrule
\end{tabular}
}
\caption{Reconstruction performance on ImageNet-1k with a resolution of $128\times 128$. All models are trained using images downsampled into $16 \times 16$ tokens. $\dagger$ Results are reproduced using the codebook size of $[8,8,8,5,5,5]$ to approximately match $65,536$. 
$+$ Following VQGAN-LC, we extract CLIP features with the codebook frozen. The codebook utilization is calculated as the fraction of the codes that are activated at least once when encoding the validation set.} 
\label{tab:image_fid}
\end{table*}

\subsection{Efficiency Analysis}

SimVQ demonstrates greater efficiency than vanilla VQ due to its asymmetric training strategy, wherein the codebook $\bm{C}$ remains static and only the linear projection $\bm{W}$ is optimized. This approach results in a significant reduction in memory usage during the gradient backpropagation process. In vanilla VQ, the memory cost for codebook optimization is $O(Kd)$, where $K$ is the number of vectors in the codebook, and $d$ is the dimension of each vector. In our experiments, $K=65,536$ is much larger than $d=128$. As the vocabulary size increases, the memory required for backpropagation grows proportionally, significantly impacting resource consumption. In contrast, SimVQ’s memory cost for backpropagation is only $O(d^2)$ because the codebook $\bm{C}$ is fixed, and only the linear layer $\bm{W}$ is updated. This results in a constant memory requirement in backpropagation, independent of the vocabulary size. The $d \times d$ scaling becomes particularly advantageous as $K$ increases in practical applications. This structural design minimizes the computational overhead and improves training efficiency, especially when dealing with large vocabularies.

\section{Experiments}
To assess the efficacy and versatility of the proposed SimVQ, we conduct experiments across both image and audio modalities. Subsequently, we analyze the learned linear layer to investigate the latent basis. The experimental configurations are listed in Appendix \ref{appendix:config}.

\subsection{Vision Modality}

\subsubsection{Baselines}
Our baseline selection focuses on methods that enhance codebook utilization through architectural designs. We include VQGAN-FC \cite{yu2022vectorquantized}, FSQ \cite{mentzer2024finite}, LFQ \cite{yu2024language} and VQGAN-LC-CLIP \cite{Zhu2024ScalingTC} as our primary baselines, as they represent the current state-of-the-art in improving reconstruction performance through codebook architecture innovations. Traditional VQ variants such as RVQ \cite{Lee2022AutoregressiveIG} and MoVQ \cite{Zheng2022HighQualityPI} address fundamentally different technical challenges - they focus on training better discrete representations, rather than enhancing codebook utilization to improve reconstruction performance like our work. Another important research direction explores optimization-based solutions, where methods like stochastic quantization \cite{pmlr-v162-takida22a}, distribution penalty \cite{VQWasserstein,Xiao2023SCVAESC}, and codebook reset \cite{zheng2023online} tackle codebook utilization through sophisticated training strategies. Given our focus on architectural innovations in codebook design, we evaluate SimVQ against methods that share this technical foundation for a meaningful assessment of our contributions.

\subsubsection{Implementation Details}
To rigorously evaluate the proposed SimVQ, we reproduce all the VQ models listed in Tab. \ref{tab:image_fid} using the same architecture of VQGAN \cite{Esser_2021_CVPR} with the quantization layer different only.
Among the baselines, for VQGAN-FC \cite{yu2022vectorquantized}, we follow the original setting to reduce the dimension of the latent space to $8$ followed by $l_2$ normalization to improve codebook utilization. For FSQ \cite{mentzer2024finite}, we adopt a codebook size of $[8,8,8,5,5,5,]$ as recommended, to approximately match the default codebook size. For VQGAN-LC \cite{Zhu2024ScalingTC}, we follow them and leverage an external pre-trained CLIP model to extract features of the training dataset in advance for a well-defined latent space. All models are trained on the ImageNet \cite{5206848} dataset for 50 epochs with a batch size of 256. Input images are processed at a resolution of $128\times 128$ pixels and downsampled by a factor of $8$, yielding a feature map of $16\times 16 \times 128$, where $128$ is the dimension of the latent space. We set the default codebook size to a large number of $2^{16}=65536$ rather than the traditional number $8192$ to highlight the representation collapse problem. Performance is evaluated using rFID, LPIPS, PSNR, and SSIM metrics on the ImageNet validation set.

\begin{table*}[t]
\centering
\resizebox{1.85\columnwidth}{!}{
\begin{tabular}{lccccccc}
\toprule
 Method & Bandwidth & Util$\uparrow$ & UTMOS$\uparrow$ & PESQ$\uparrow$ & STOI$\uparrow$ & V/UV F1$\uparrow$ \\
 \midrule
GT & - & - & 4.06/3.48 & - & - & - \\
EnCodec \cite{efossez2023high} & 3.0kbps & - & 2.31/2.09 & 2.05/2.05 & 0.90/0.88 & 0.92/0.89 \\
Vocos \cite{siuzdak2024vocos} & 3.0kbps & - & 3.53/3.06 & 2.40/2.19 & 0.92/0.90 & 0.94/0.91 \\
SpeechTokenizer \cite{zhang2024speechtokenizer} & 3.0kbps & - & 3.56/3.02 & 1.93/1.74 & 0.88/0.84 & 0.93/0.89 \\
\midrule
WavTokenizer \cite{ji2024wavtokenizer} & 0.9kbps & 100/100\% & 3.74/3.43$^*$ & 2.01/2.26$^*$ & 0.89/0.89$^*$ & 0.92/0.92$^*$ \\
SimVQ (ours) & 0.9kbps & 100.0/100.0\% & \textbf{4.00/3.51} & \textbf{2.33}/2.08 & \textbf{0.91}/0.88 & \textbf{0.94}/0.91 \\
\midrule
WavTokenizer \cite{ji2024wavtokenizer} & 0.975kbps & 68/-\% & 4.02$^*$/- & 2.39$^*$/- & 0.92$^*$/- & 0.94$^*$/- \\
WavTokenizer \cite{ji2024wavtokenizer} & 1.05kbps & 27/-\% & 4.00$^*$/- & 2.36$^*$/- & 0.81$^*$/- & 0.94$^*$/- \\
SimVQ (ours) & 0.975kbps & 99.4/99.4\% & 4.03/3.52 & 2.42/2.15 & 0.92/0.88 & 0.94/0.92 \\
SimVQ (ours) & 1.2kbps & 99.4/99.0\% & 4.03/3.52 & 2.54/2.26 & 0.93/0.90 & 0.94/0.92 \\
SimVQ (ours) & 1.35kbps & 95.6/94.7\% & \textbf{4.03/3.53} & \textbf{2.61/2.31} & \textbf{0.93/0.90} & \textbf{0.95/0.93} \\
\bottomrule
\end{tabular}
}
\caption{Reconstruction performance on LibriTTS test-clean/test-other dataset. $*$ WavTokenizer is trained with a window size of 3 seconds. The bandwidth of 0.9kbps, 0.975kbps, 1.2kbps, 1.35kbps means the codebook size of 4096, 8192, 65536, 262144 respectively.}
\label{tab:audio_utmos}
\end{table*}

\subsubsection{Main Results}
Tab. \ref{tab:image_fid} presents the reconstruction performance of various VQ models on image data. We make three key observations: 1) Traditional VQGAN models utilize only a very small subset of the codebook, with a utilization rate of just $1.4\%$. Although VQGAN-EMA is proposed to improve codebook utilization, especially when the codebook size scales up to $65k$, it still suffers from severe representation collapse. 2) Recently proposed methods, such as LFQ, FSQ, and VQGAN-FC, effectively improve codebook utilization to $100\%$. However, these methods require reducing the latent space to a very low dimension. For example, applying VQGAN-FC to the standard latent dimension of $128$ results in severe representation collapse and degraded reconstruction performance. Additionally, these models face limitations in model capacity due to the low-dimensional latent space. While they achieve full codebook utilization, their reconstruction quality on rFID score lags significantly behind SimVQ. 3) VQGAN-LC-CLIP leverages an external pre-trained CLIP model to provide a well-defined latent space. However, VQGAN-LC relies on CLIP features pre-trained on much larger datasets than ImageNet, which introduces generalization issues and a lower performance ceiling (degradation issue in Tab. \ref{tab:ablation_codebook}). In contrast, SimVQ can be applied to a wide range of data types and achieves superior performance (rFID $2.40$ vs. $2.24$) without the limitations imposed by a pre-trained feature extraction model.

\begin{table}[t]
\centering
\resizebox{1.0\columnwidth}{!}{
\begin{tabular}{lcccccc}
\toprule
 Method & Codebook Size & Util$\uparrow$ & rFID$\downarrow$ & LPIPS$\downarrow$ & PSNR$\uparrow$ & SSIM$\uparrow$ \\
\midrule
VQGAN-LC-CLIP$^{\dagger}$ \cite{Zhu2024ScalingTC} & 50,000 & $99.9\%$ & 2.75 & 0.13 & 23.8 & 58.4 \\
VQGAN-LC-CLIP$^{\dagger}$ \cite{Zhu2024ScalingTC} & 100,000 & $99.9\%$ & \underline{2.62}& 0.12& 23.8 & 58.9 \\
VQGAN-LC-CLIP$^{\dagger}$ \cite{Zhu2024ScalingTC} & 200,000 & $99.8\%$ & \underline{2.66}& 0.12& 23.9 & 59.2 \\
 \midrule
SimVQ & 1,024 & 100.0\% & 3.67 & 0.16 & 22.34 & 70.8 \\
SimVQ & 8,192 & 100.0\% & 2.98 & 0.14 & 23.23 & 74.7 \\
SimVQ & 65,536 & 100.0\% & 2.24 & 0.12 & 24.15 & 78.4 \\
SimVQ & 262,144 & 100.0\% & \textbf{1.99} & \textbf{0.11} & \textbf{24.68} & \textbf{80.3} \\
\bottomrule
\end{tabular}
}
\caption{Ablation study on the effect of various codebook sizes on ImageNet at a resolution of $128 \times 128$. $\dagger$ We directly copy the reported results of VQGAN-LC from the original paper on ImageNet $256\times 256$ resolution.}
\label{tab:ablation_codebook}
\end{table}

\begin{table}[t]
\centering
\resizebox{1.0\columnwidth}{!}{
\begin{tabular}{ccccccc}
\toprule
 Initialization & Trainable & Util$\uparrow$ & rFID$\downarrow$ & LPIPS$\downarrow$ & PSNR$\uparrow$ & SSIM$\uparrow$ \\
 \midrule
Gaussian & Yes & 100.0\% & 2.31 & 0.12 & 24.04 & 77.2 \\
Uniform & No & 100.0\% & 2.31 & 0.12 & 24.15 & 78.4 \\
Gaussian & No & 100.0\% & \textbf{2.24} & \textbf{0.12} & \textbf{24.15} & \textbf{78.4} \\
\bottomrule
\end{tabular}
}
\caption{Ablation study of codebook optimization strategy.}
\label{tab:ablation_optim}
\end{table}

\subsubsection{Ablation Study}

\paragraph{On the Codebook Size} 
In Tab. \ref{tab:ablation_codebook}, we explore the impact of different codebook sizes, ranging from $1k$ to $262k$, which is typically the level of LLM's vocabulary size. SimVQ consistently improves performance as the codebook size increases. For instance, the rFID score decreases to $1.99$, and SSIM surpasses $80.0$. In contrast, while VQGAN-LC-CLIP can keep high codebook utilization as increasing codebook size, it encounters performance degradation, with the rFID score worsening from $2.62$ to $2.66$ when the codebook size is increased from $100,000$ to $200,000$.

\paragraph{On the Codebook Optimization Strategy}
We investigate codebook initialization and the training of the linear layer in Tab. \ref{tab:ablation_optim}. Our findings are as follows: 1) The codebook is robust to different initialization strategies, yielding similar results with both Gaussian and uniform initialization. 2) When the codebook is updated during training, SimVQ continues to address the representation collapse issue, though with a slight degradation in performance.

\subsection{Audio Modality}

\subsubsection{Baselines and Implementation Details}

We use the LibriTTS dataset \cite{zen2019libritts} for audio-based VQ model training.  The baselines such as Encodec \cite{efossez2023high}, Vocos \cite{siuzdak2024vocos}, and SpeechTokenizer \cite{zhang2024speechtokenizer} are based on residual vector quantization method. Our SimVQ model adopts the same architecture as WavTokenizer \cite{ji2024wavtokenizer} with the only modification being the replacement of their EMA codebook with our one linear layer reparameterization method. We train SimVQ on LibriTTS-580h for 50 epochs with a batch size of 64. Note that WavTokenizer is trained with a 3-second window size for optimal performance, we train SimVQ using a 1-second window to accelerate training. For objective evaluation of the reconstructed audio, we follow Vocos \cite{siuzdak2024vocos} and employ metrics such as UTMOS \cite{Saeki2022UTMOSUS}, PESQ \cite{941023}, STOI, and the F1 score for voiced/unvoiced classification (V/UV F1). UTMOS is particularly valuable as it produces scores highly correlated with human evaluations.

\subsubsection{Main Results}
Tab. \ref{tab:audio_utmos} presents the reconstruction performance of various VQ models on audio data. Baseline models using residual vector quantization perform significantly worse than SimVQ, even when utilizing much larger bandwidths. Despite using the same architecture as WavTokenizer, our model, which replaces the quantization layer with SimVQ, achieves superior performance with a 1-second window size and maintains nearly $100\%$ codebook utilization when scaling up to a size of 262,144. The consistent performance of the SimVQ model across both image and audio data demonstrates that SimVQ is a general method for addressing the representation collapse problem in VQ models and can be effectively applied across multiple modalities.

\subsection{Analysis}

\begin{figure}[t]
    \centering
    \begin{minipage}[b]{0.49\columnwidth}
        \centering
        \includegraphics[width=\columnwidth]{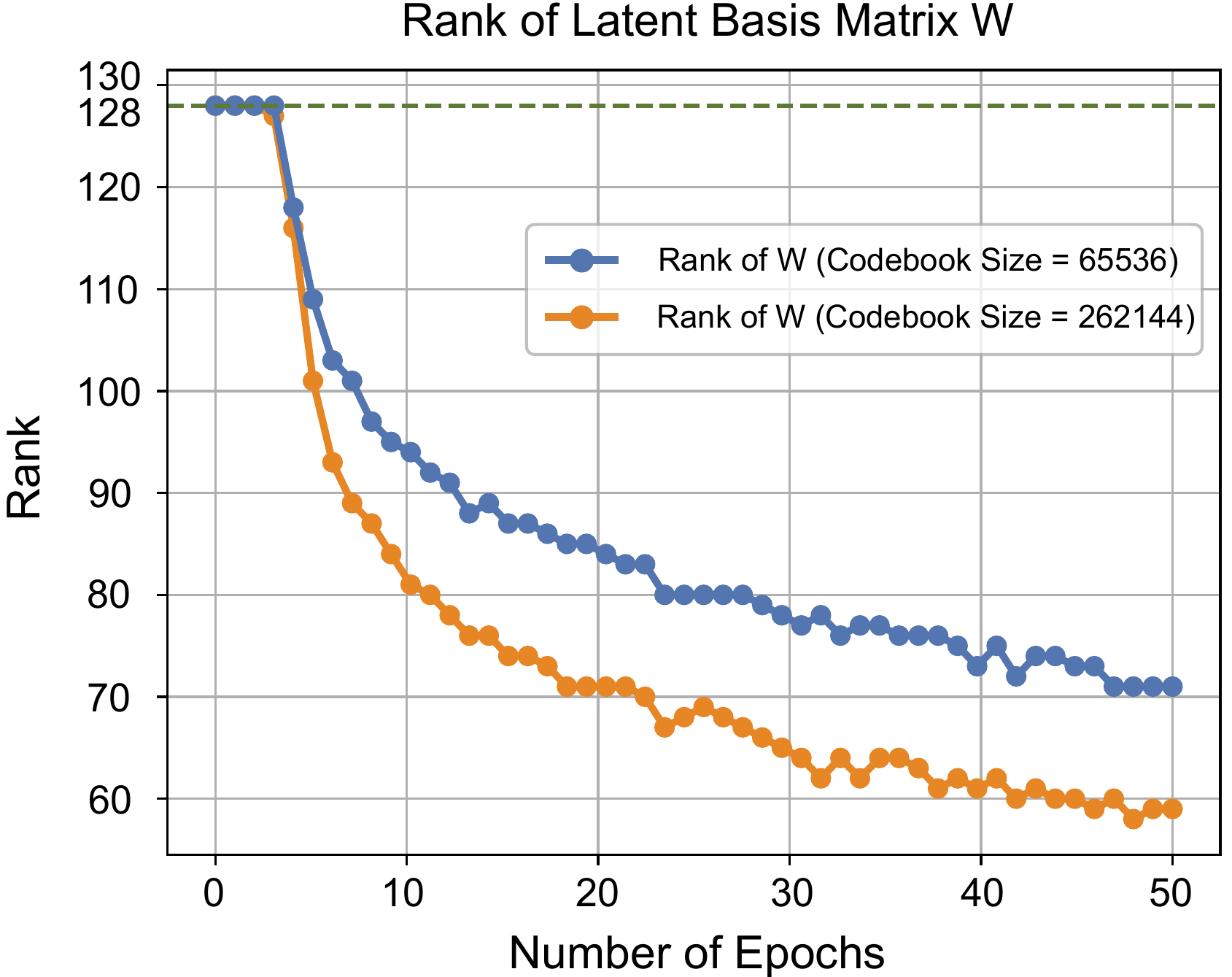}
    \end{minipage}
    \begin{minipage}[b]{0.49\columnwidth}
        \centering
        \includegraphics[width=\columnwidth]{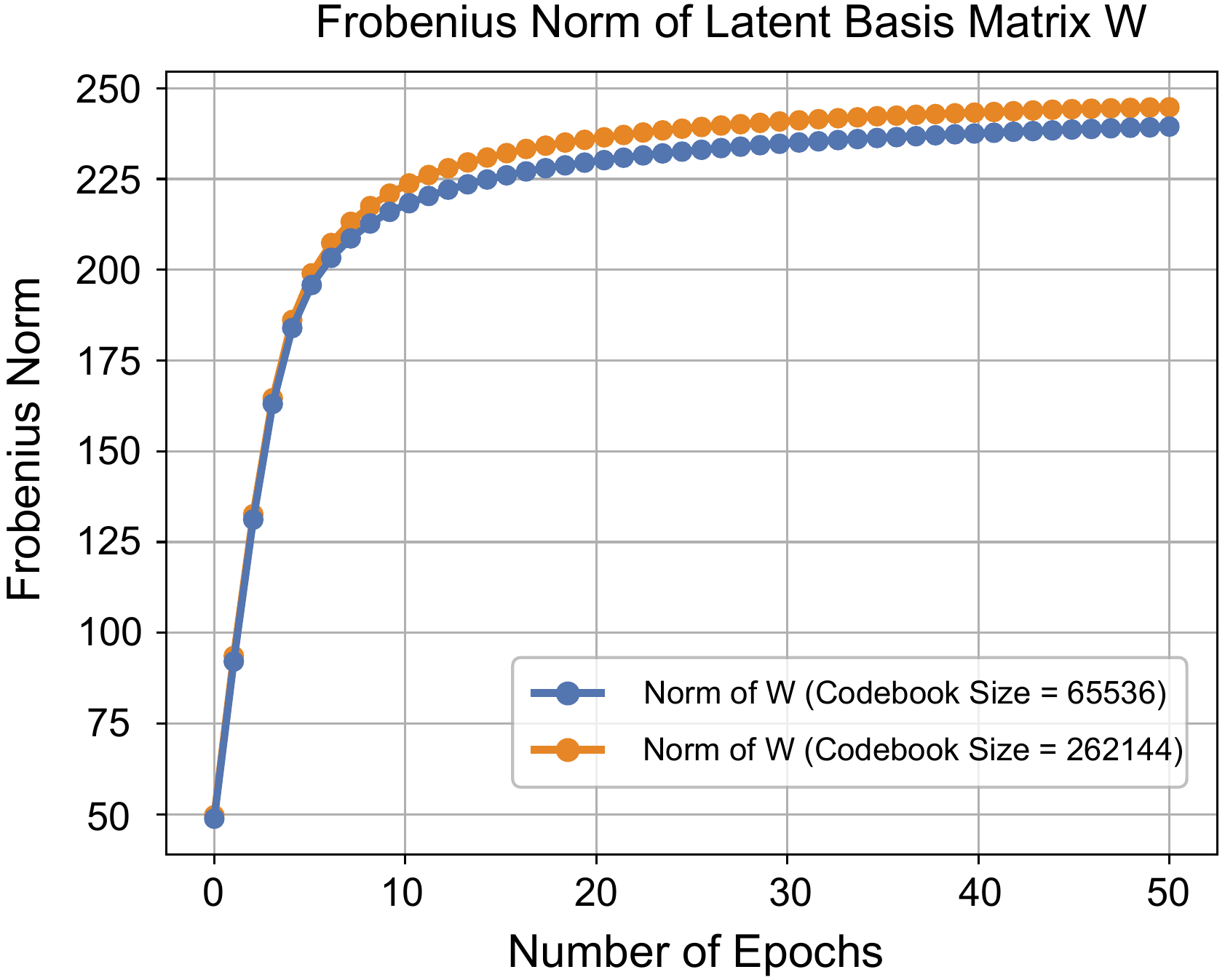}
    \end{minipage}
    \caption{(a):(left) The rank of latent basis matrix $\bm{W}$ over training epochs. (b):(right) The Frobenius norm of latent basis matrix $\bm{W}$ over training epochs.}
    \label{fig:rank_norm}
\end{figure}

In Fig. \ref{fig:rank_norm}(a), we plot the rank of the latent basis matrix over training epochs. Notably, SimVQ demonstrates the ability to adaptively adjust the rank of the latent space. Specifically, when the codebook size increases from $65k$ to $262k$, the rank of the latent basis matrix decreases more rapidly and converges to a lower value. This observation suggests that a larger codebook can effectively alleviate the pressure on the latent space dimensionality, allowing the model to learn to represent data more efficiently. Additionally, despite the rank decreasing to a lower-rank space, SimVQ maintains $100\%$ codebook utilization, highlighting its superiority over VQGAN-FC, which struggles when increasing the latent dimension from 8 to 128.
We also calculate the Frobenius norm of the latent basis matrix, as shown in Fig. \ref{fig:rank_norm}. The norm of a codebook size of $262k$ is slightly larger than for $65k$, indicating that a larger codebook can span a broader area in the linear space.
For a comprehensive evaluation, we also provide the reconstruction loss curve on the ImageNet validation dataset in Appendix \ref{appendix:loss}. The results consistently show that SimVQ achieves improved performance, further validating the effectiveness of our approach.

\begin{figure}[t]
    \centering
    \includegraphics[width=0.8\columnwidth]{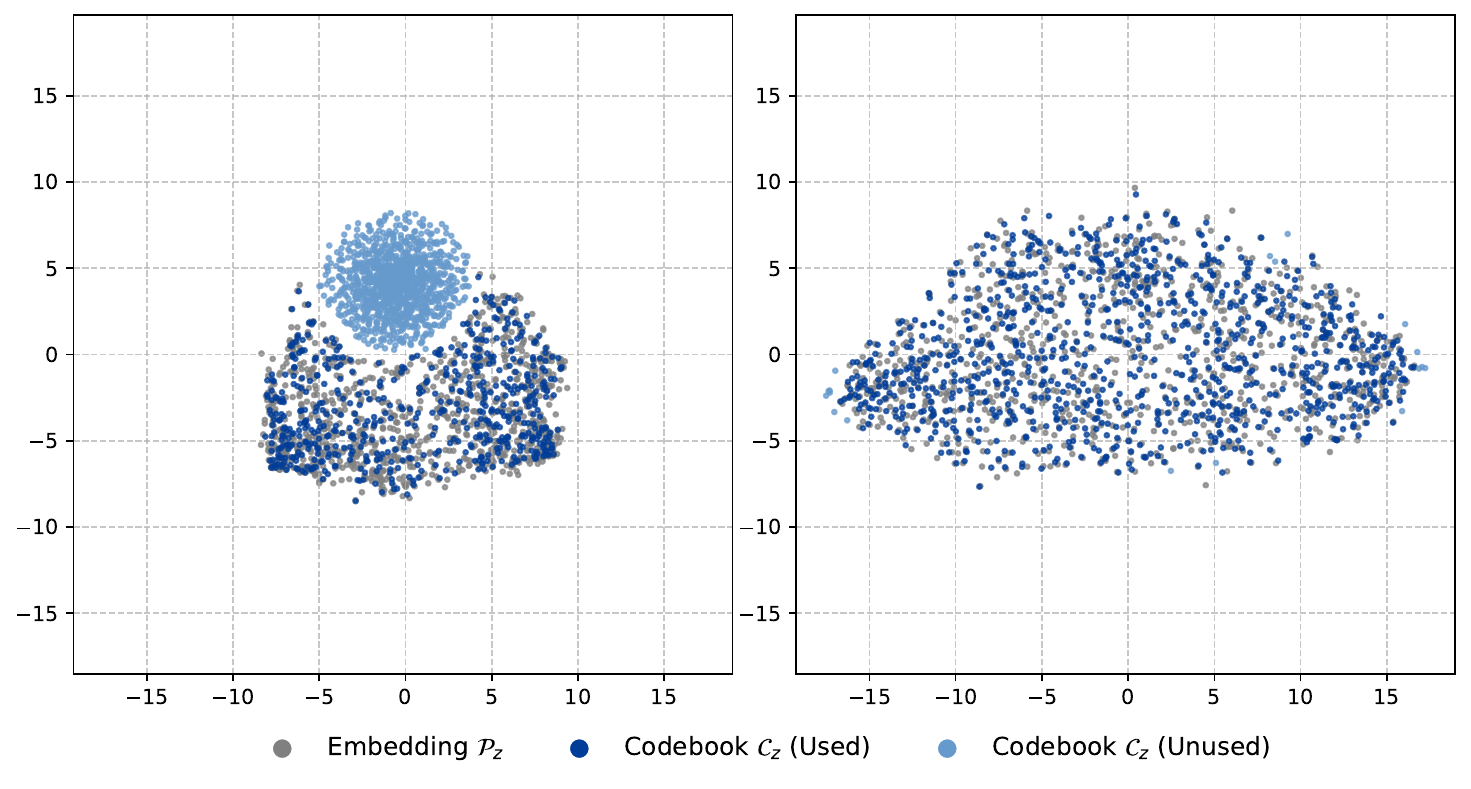}
    \caption{Visualization of the divergence between the encoder features and codebook embeddings on a random subset of ImageNet validation dataset. The left figure is vanilla VQ model and the right one is SimVQ.}
    \label{fig:visualization}
\end{figure}

\subsection{Qualitative Evaluation}
We visualize the distribution of encoder features and codebook embeddings in Fig. \ref{fig:visualization} and the frequency in Appendix \ref{appendix:freq}. Compared to vanilla VQ models that most codebook vectors are not unused, SimVQ can update the whole codebook and distribute the features evenly in the whole space.
We qualitatively compare the reconstruction quality of both images and audio in Appendix \ref{appendix:cases}. SimVQ can presreve more details with an enlarged codebook size, such as ``eyes'' and ``text'', which are challenging for vanilla VQ models.

\section{Discussion}

\paragraph{VQ Performance and Generative Models}

Many generative models \cite{Rombach_2022_CVPR,yu2022scaling,sun2024autoregressive,team2024chameleon} utilize VQ models as a ``tokenizer'' to obtain discrete tokens. While it is intuitive that VQ reconstruction quality should impact generation performance, recent studies \cite{zhu2024stabilize,yu2024language} have revealed that the relationship is more nuanced: improved VQ reconstruction metrics do not necessarily translate to better generative outcomes. This complex relationship suggests that evaluating VQ models primarily through downstream generation tasks may not provide the most insightful assessment of their fundamental properties. Therefore, we focus on addressing the critical issue of representation collapse in VQ models, aiming to advance our understanding of their core representation learning mechanisms.

\section{Conclusion}
In this paper, we explore the representation collapse problem in VQ models. We conduct a theoretical analysis of the optimization process in VQ models and propose a simple yet effective method, SimVQ, to address this issue. Our method addresses the representation collapse by jointly optimizing the latent space through a linear transformation with one linear layer. Experimental results demonstrate that SimVQ outperforms previous approaches on both image and audio datasets, highlighting its broad applicability across diverse machine learning tasks.

\section*{Acknowledgement}

This research was supported by the National Natural Science Foundation of China (Grant No.62276245).
{
    \small
    \bibliographystyle{ieeenat_fullname}
    \bibliography{main}

\begin{thebibliography}{44}
\providecommand{\natexlab}[1]{#1}
\providecommand{\url}[1]{\texttt{#1}}
\expandafter\ifx\csname urlstyle\endcsname\relax
  \providecommand{\doi}[1]{doi: #1}\else
  \providecommand{\doi}{doi: \begingroup \urlstyle{rm}\Url}\fi

\bibitem[Achiam et~al.(2023)Achiam, Adler, Agarwal, Ahmad, Akkaya, Aleman, Almeida, Altenschmidt, Altman, Anadkat, et~al.]{achiam2023gpt}
Josh Achiam, Steven Adler, Sandhini Agarwal, Lama Ahmad, Ilge Akkaya, Florencia~Leoni Aleman, Diogo Almeida, Janko Altenschmidt, Sam Altman, Shyamal Anadkat, et~al.
\newblock Gpt-4 technical report.
\newblock \emph{arXiv preprint arXiv:2303.08774}, 2023.

\bibitem[Baevski et~al.(2020)Baevski, Zhou, Mohamed, and Auli]{NEURIPS2020_92d1e1eb}
Alexei Baevski, Yuhao Zhou, Abdelrahman Mohamed, and Michael Auli.
\newblock wav2vec 2.0: A framework for self-supervised learning of speech representations.
\newblock In \emph{Advances in Neural Information Processing Systems}, pages 12449--12460. Curran Associates, Inc., 2020.

\bibitem[Bengio et~al.(2013)Bengio, L{\'e}onard, and Courville]{bengio2013estimating}
Yoshua Bengio, Nicholas L{\'e}onard, and Aaron Courville.
\newblock Estimating or propagating gradients through stochastic neurons for conditional computation.
\newblock \emph{arXiv preprint arXiv:1308.3432}, 2013.

\bibitem[Borsos et~al.(2023)Borsos, Marinier, Vincent, Kharitonov, Pietquin, Sharifi, Roblek, Teboul, Grangier, Tagliasacchi, and Zeghidour]{10158503}
Zalán Borsos, Raphaël Marinier, Damien Vincent, Eugene Kharitonov, Olivier Pietquin, Matt Sharifi, Dominik Roblek, Olivier Teboul, David Grangier, Marco Tagliasacchi, and Neil Zeghidour.
\newblock Audiolm: A language modeling approach to audio generation.
\newblock \emph{IEEE/ACM Transactions on Audio, Speech, and Language Processing}, 31:\penalty0 2523--2533, 2023.

\bibitem[Bruce et~al.(2024)Bruce, Dennis, Edwards, Parker-Holder, Shi, Hughes, Lai, Mavalankar, Steigerwald, Apps, Aytar, Bechtle, Behbahani, Chan, Heess, Gonzalez, Osindero, Ozair, Reed, Zhang, Zolna, Clune, Freitas, Singh, and Rockt\"{a}schel]{pmlr-v235-bruce24a}
Jake Bruce, Michael~D Dennis, Ashley Edwards, Jack Parker-Holder, Yuge Shi, Edward Hughes, Matthew Lai, Aditi Mavalankar, Richie Steigerwald, Chris Apps, Yusuf Aytar, Sarah Maria~Elisabeth Bechtle, Feryal Behbahani, Stephanie~C.Y. Chan, Nicolas Heess, Lucy Gonzalez, Simon Osindero, Sherjil Ozair, Scott Reed, Jingwei Zhang, Konrad Zolna, Jeff Clune, Nando~De Freitas, Satinder Singh, and Tim Rockt\"{a}schel.
\newblock Genie: Generative interactive environments.
\newblock In \emph{Proceedings of the 41st International Conference on Machine Learning}, pages 4603--4623. PMLR, 2024.

\bibitem[D{\'e}fossez et~al.(2023)D{\'e}fossez, Copet, Synnaeve, and Adi]{efossez2023high}
Alexandre D{\'e}fossez, Jade Copet, Gabriel Synnaeve, and Yossi Adi.
\newblock High fidelity neural audio compression.
\newblock \emph{Transactions on Machine Learning Research}, 2023.
\newblock Featured Certification, Reproducibility Certification.

\bibitem[Deng et~al.(2009)Deng, Dong, Socher, Li, Li, and Fei-Fei]{5206848}
Jia Deng, Wei Dong, Richard Socher, Li-Jia Li, Kai Li, and Li Fei-Fei.
\newblock Imagenet: A large-scale hierarchical image database.
\newblock In \emph{2009 IEEE Conference on Computer Vision and Pattern Recognition}, pages 248--255, 2009.

\bibitem[Dubey et~al.(2024)Dubey, Jauhri, Pandey, Kadian, Al-Dahle, Letman, Mathur, Schelten, Yang, Fan, et~al.]{dubey2024llama}
Abhimanyu Dubey, Abhinav Jauhri, Abhinav Pandey, Abhishek Kadian, Ahmad Al-Dahle, Aiesha Letman, Akhil Mathur, Alan Schelten, Amy Yang, Angela Fan, et~al.
\newblock The llama 3 herd of models.
\newblock \emph{arXiv preprint arXiv:2407.21783}, 2024.

\bibitem[Esser et~al.(2021)Esser, Rombach, and Ommer]{Esser_2021_CVPR}
Patrick Esser, Robin Rombach, and Bjorn Ommer.
\newblock Taming transformers for high-resolution image synthesis.
\newblock In \emph{Proceedings of the IEEE/CVF Conference on Computer Vision and Pattern Recognition (CVPR)}, pages 12873--12883, 2021.

\bibitem[Gao et~al.(2024)Gao, Tan, Wang, Huang, Wu, and Li]{gao2024foldtoken}
Zhangyang Gao, Cheng Tan, Jue Wang, Yufei Huang, Lirong Wu, and Stan~Z Li.
\newblock Foldtoken: Learning protein language via vector quantization and beyond.
\newblock \emph{arXiv preprint arXiv:2403.09673}, 2024.

\bibitem[Huh et~al.(2023)Huh, Cheung, Agrawal, and Isola]{pmlr-v202-huh23a}
Minyoung Huh, Brian Cheung, Pulkit Agrawal, and Phillip Isola.
\newblock Straightening out the straight-through estimator: Overcoming optimization challenges in vector quantized networks.
\newblock In \emph{Proceedings of the 40th International Conference on Machine Learning}, pages 14096--14113. PMLR, 2023.

\bibitem[Jang et~al.(2017)Jang, Gu, and Poole]{DBLP:conf/iclr/JangGP17}
Eric Jang, Shixiang Gu, and Ben Poole.
\newblock Categorical reparameterization with gumbel-softmax.
\newblock In \emph{5th International Conference on Learning Representations, {ICLR} 2017, Toulon, France, April 24-26, 2017, Conference Track Proceedings}. OpenReview.net, 2017.

\bibitem[Ji et~al.(2024)Ji, Jiang, Cheng, Chen, Fang, Zuo, Yang, Li, Zhang, Yang, et~al.]{ji2024wavtokenizer}
Shengpeng Ji, Ziyue Jiang, Xize Cheng, Yifu Chen, Minghui Fang, Jialong Zuo, Qian Yang, Ruiqi Li, Ziang Zhang, Xiaoda Yang, et~al.
\newblock Wavtokenizer: an efficient acoustic discrete codec tokenizer for audio language modeling.
\newblock \emph{arXiv preprint arXiv:2408.16532}, 2024.

\bibitem[Kingma and Welling(2013)]{Kingma2013AutoEncodingVB}
Diederik~P. Kingma and Max Welling.
\newblock Auto-encoding variational bayes.
\newblock \emph{CoRR}, abs/1312.6114, 2013.

\bibitem[Lee et~al.(2022)Lee, Kim, Kim, Cho, and Han]{Lee2022AutoregressiveIG}
Doyup Lee, Chiheon Kim, Saehoon Kim, Minsu Cho, and Wook-Shin Han.
\newblock Autoregressive image generation using residual quantization.
\newblock \emph{2022 IEEE/CVF Conference on Computer Vision and Pattern Recognition (CVPR)}, pages 11513--11522, 2022.

\bibitem[Mentzer et~al.(2024)Mentzer, Minnen, Agustsson, and Tschannen]{mentzer2024finite}
Fabian Mentzer, David Minnen, Eirikur Agustsson, and Michael Tschannen.
\newblock Finite scalar quantization: {VQ}-{VAE} made simple.
\newblock In \emph{The Twelfth International Conference on Learning Representations}, 2024.

\bibitem[Radford et~al.(2021)Radford, Kim, Hallacy, Ramesh, Goh, Agarwal, Sastry, Askell, Mishkin, Clark, Krueger, and Sutskever]{pmlr-v139-radford21a}
Alec Radford, Jong~Wook Kim, Chris Hallacy, Aditya Ramesh, Gabriel Goh, Sandhini Agarwal, Girish Sastry, Amanda Askell, Pamela Mishkin, Jack Clark, Gretchen Krueger, and Ilya Sutskever.
\newblock Learning transferable visual models from natural language supervision.
\newblock In \emph{Proceedings of the 38th International Conference on Machine Learning}, pages 8748--8763. PMLR, 2021.

\bibitem[Ramesh et~al.(2021)Ramesh, Pavlov, Goh, Gray, Voss, Radford, Chen, and Sutskever]{pmlr-v139-ramesh21a}
Aditya Ramesh, Mikhail Pavlov, Gabriel Goh, Scott Gray, Chelsea Voss, Alec Radford, Mark Chen, and Ilya Sutskever.
\newblock Zero-shot text-to-image generation.
\newblock In \emph{Proceedings of the 38th International Conference on Machine Learning}, pages 8821--8831. PMLR, 2021.

\bibitem[Razavi et~al.(2019)Razavi, van~den Oord, and Vinyals]{NEURIPS2019_5f8e2fa1}
Ali Razavi, Aaron van~den Oord, and Oriol Vinyals.
\newblock Generating diverse high-fidelity images with vq-vae-2.
\newblock In \emph{Advances in Neural Information Processing Systems}. Curran Associates, Inc., 2019.

\bibitem[Rix et~al.(2001)Rix, Beerends, Hollier, and Hekstra]{941023}
A.W. Rix, J.G. Beerends, M.P. Hollier, and A.P. Hekstra.
\newblock Perceptual evaluation of speech quality (pesq)-a new method for speech quality assessment of telephone networks and codecs.
\newblock In \emph{2001 IEEE International Conference on Acoustics, Speech, and Signal Processing. Proceedings (Cat. No.01CH37221)}, pages 749--752 vol.2, 2001.

\bibitem[Rombach et~al.(2022)Rombach, Blattmann, Lorenz, Esser, and Ommer]{Rombach_2022_CVPR}
Robin Rombach, Andreas Blattmann, Dominik Lorenz, Patrick Esser, and Bj\"orn Ommer.
\newblock High-resolution image synthesis with latent diffusion models.
\newblock In \emph{Proceedings of the IEEE/CVF Conference on Computer Vision and Pattern Recognition (CVPR)}, pages 10684--10695, 2022.

\bibitem[Roy et~al.(2018)Roy, Vaswani, Neelakantan, and Parmar]{roy2018theory}
Aurko Roy, Ashish Vaswani, Arvind Neelakantan, and Niki Parmar.
\newblock Theory and experiments on vector quantized autoencoders.
\newblock \emph{arXiv preprint arXiv:1805.11063}, 2018.

\bibitem[Saeki et~al.(2022)Saeki, Xin, Nakata, Koriyama, Takamichi, and Saruwatari]{Saeki2022UTMOSUS}
Takaaki Saeki, Detai Xin, Wataru Nakata, Tomoki Koriyama, Shinnosuke Takamichi, and Hiroshi Saruwatari.
\newblock Utmos: Utokyo-sarulab system for voicemos challenge 2022.
\newblock \emph{ArXiv}, abs/2204.02152, 2022.

\bibitem[Siuzdak(2024)]{siuzdak2024vocos}
Hubert Siuzdak.
\newblock Vocos: Closing the gap between time-domain and fourier-based neural vocoders for high-quality audio synthesis.
\newblock In \emph{The Twelfth International Conference on Learning Representations}, 2024.

\bibitem[Sun et~al.(2024)Sun, Jiang, Chen, Zhang, Peng, Luo, and Yuan]{sun2024autoregressive}
Peize Sun, Yi Jiang, Shoufa Chen, Shilong Zhang, Bingyue Peng, Ping Luo, and Zehuan Yuan.
\newblock Autoregressive model beats diffusion: Llama for scalable image generation.
\newblock \emph{arXiv preprint arXiv:2406.06525}, 2024.

\bibitem[Takida et~al.(2022)Takida, Shibuya, Liao, Lai, Ohmura, Uesaka, Murata, Takahashi, Kumakura, and Mitsufuji]{pmlr-v162-takida22a}
Yuhta Takida, Takashi Shibuya, Weihsiang Liao, Chieh-Hsin Lai, Junki Ohmura, Toshimitsu Uesaka, Naoki Murata, Shusuke Takahashi, Toshiyuki Kumakura, and Yuki Mitsufuji.
\newblock {SQ}-{VAE}: Variational {B}ayes on discrete representation with self-annealed stochastic quantization.
\newblock In \emph{Proceedings of the 39th International Conference on Machine Learning}, pages 20987--21012. PMLR, 2022.

\bibitem[Tao et~al.(2024)Tao, Liu, Dou, Muennighoff, Wan, Luo, Lin, and Wong]{tao2024scaling}
Chaofan Tao, Qian Liu, Longxu Dou, Niklas Muennighoff, Zhongwei Wan, Ping Luo, Min Lin, and Ngai Wong.
\newblock Scaling laws with vocabulary: Larger models deserve larger vocabularies.
\newblock \emph{arXiv preprint arXiv:2407.13623}, 2024.

\bibitem[Team(2024)]{team2024chameleon}
Chameleon Team.
\newblock Chameleon: Mixed-modal early-fusion foundation models.
\newblock \emph{arXiv preprint arXiv:2405.09818}, 2024.

\bibitem[van~den Oord et~al.(2017)van~den Oord, Vinyals, and kavukcuoglu]{NIPS2017_7a98af17}
Aaron van~den Oord, Oriol Vinyals, and koray kavukcuoglu.
\newblock Neural discrete representation learning.
\newblock In \emph{Advances in Neural Information Processing Systems}. Curran Associates, Inc., 2017.

\bibitem[Vuong et~al.(2023)Vuong, Le, Zhao, Zheng, Harandi, Cai, and Phung]{VQWasserstein}
Tung-Long Vuong, Trung Le, He Zhao, Chuanxia Zheng, Mehrtash Harandi, Jianfei Cai, and Dinh Phung.
\newblock Vector quantized wasserstein auto-encoder.
\newblock In \emph{Proceedings of the 40th International Conference on Machine Learning}. JMLR.org, 2023.

\bibitem[Wang et~al.(2023)Wang, Chen, Wu, Zhang, Zhou, Liu, Chen, Liu, Wang, Li, et~al.]{wang2023neural}
Chengyi Wang, Sanyuan Chen, Yu Wu, Ziqiang Zhang, Long Zhou, Shujie Liu, Zhuo Chen, Yanqing Liu, Huaming Wang, Jinyu Li, et~al.
\newblock Neural codec language models are zero-shot text to speech synthesizers.
\newblock \emph{arXiv preprint arXiv:2301.02111}, 2023.

\bibitem[Xiao et~al.(2023)Xiao, Qiu, and Sotiras]{Xiao2023SCVAESC}
Pan Xiao, Peijie Qiu, and Aristeidis Sotiras.
\newblock Sc-vae: Sparse coding-based variational autoencoder.
\newblock \emph{ArXiv}, abs/2303.16666, 2023.

\bibitem[Yu et~al.(2022{\natexlab{a}})Yu, Li, Koh, Zhang, Pang, Qin, Ku, Xu, Baldridge, and Wu]{yu2022vectorquantized}
Jiahui Yu, Xin Li, Jing~Yu Koh, Han Zhang, Ruoming Pang, James Qin, Alexander Ku, Yuanzhong Xu, Jason Baldridge, and Yonghui Wu.
\newblock Vector-quantized image modeling with improved {VQGAN}.
\newblock In \emph{International Conference on Learning Representations}, 2022{\natexlab{a}}.

\bibitem[Yu et~al.(2022{\natexlab{b}})Yu, Xu, Koh, Luong, Baid, Wang, Vasudevan, Ku, Yang, Ayan, Hutchinson, Han, Parekh, Li, Zhang, Baldridge, and Wu]{yu2022scaling}
Jiahui Yu, Yuanzhong Xu, Jing~Yu Koh, Thang Luong, Gunjan Baid, Zirui Wang, Vijay Vasudevan, Alexander Ku, Yinfei Yang, Burcu~Karagol Ayan, Ben Hutchinson, Wei Han, Zarana Parekh, Xin Li, Han Zhang, Jason Baldridge, and Yonghui Wu.
\newblock Scaling autoregressive models for content-rich text-to-image generation.
\newblock \emph{Transactions on Machine Learning Research}, 2022{\natexlab{b}}.
\newblock Featured Certification.

\bibitem[Yu et~al.(2024)Yu, Lezama, Gundavarapu, Versari, Sohn, Minnen, Cheng, Gupta, Gu, Hauptmann, Gong, Yang, Essa, Ross, and Jiang]{yu2024language}
Lijun Yu, Jose Lezama, Nitesh~Bharadwaj Gundavarapu, Luca Versari, Kihyuk Sohn, David Minnen, Yong Cheng, Agrim Gupta, Xiuye Gu, Alexander~G Hauptmann, Boqing Gong, Ming-Hsuan Yang, Irfan Essa, David~A Ross, and Lu Jiang.
\newblock Language model beats diffusion - tokenizer is key to visual generation.
\newblock In \emph{The Twelfth International Conference on Learning Representations}, 2024.

\bibitem[Zen et~al.(2019)Zen, Dang, Clark, Zhang, Weiss, Jia, Chen, and Wu]{zen2019libritts}
Heiga Zen, Viet Dang, Rob Clark, Yu Zhang, Ron~J Weiss, Ye Jia, Zhifeng Chen, and Yonghui Wu.
\newblock Libritts: A corpus derived from librispeech for text-to-speech.
\newblock \emph{arXiv preprint arXiv:1904.02882}, 2019.

\bibitem[Zhang et~al.(2023{\natexlab{a}})Zhang, Li, Zhang, Zhan, Wang, Zhou, and Qiu]{zhang-etal-2023-speechgpt}
Dong Zhang, Shimin Li, Xin Zhang, Jun Zhan, Pengyu Wang, Yaqian Zhou, and Xipeng Qiu.
\newblock {S}peech{GPT}: Empowering large language models with intrinsic cross-modal conversational abilities.
\newblock In \emph{Findings of the Association for Computational Linguistics: EMNLP 2023}, pages 15757--15773, Singapore, 2023{\natexlab{a}}. Association for Computational Linguistics.

\bibitem[Zhang et~al.(2023{\natexlab{b}})Zhang, Zhan, Theobalt, and Lu]{Zhang_2023_CVPR}
Jiahui Zhang, Fangneng Zhan, Christian Theobalt, and Shijian Lu.
\newblock Regularized vector quantization for tokenized image synthesis.
\newblock In \emph{Proceedings of the IEEE/CVF Conference on Computer Vision and Pattern Recognition (CVPR)}, pages 18467--18476, 2023{\natexlab{b}}.

\bibitem[Zhang et~al.(2024)Zhang, Zhang, Li, Zhou, and Qiu]{zhang2024speechtokenizer}
Xin Zhang, Dong Zhang, Shimin Li, Yaqian Zhou, and Xipeng Qiu.
\newblock Speechtokenizer: Unified speech tokenizer for speech language models.
\newblock In \emph{The Twelfth International Conference on Learning Representations}, 2024.

\bibitem[Zheng and Vedaldi(2023)]{zheng2023online}
Chuanxia Zheng and Andrea Vedaldi.
\newblock Online clustered codebook.
\newblock In \emph{Proceedings of the IEEE/CVF International Conference on Computer Vision}, pages 22798--22807, 2023.

\bibitem[Zheng et~al.(2022)Zheng, Song, Cham, Cai, Phung, and Luo]{Zheng2022HighQualityPI}
Chuanxia Zheng, Guoxian Song, Tat-Jen Cham, Jianfei Cai, Dinh~Q. Phung, and Linjie Luo.
\newblock High-quality pluralistic image completion via code shared vqgan.
\newblock \emph{ArXiv}, abs/2204.01931, 2022.

\bibitem[Zhu et~al.(2024{\natexlab{a}})Zhu, Wei, Lu, and Chen]{Zhu2024ScalingTC}
Lei Zhu, Fangyun Wei, Yanye Lu, and Dong Chen.
\newblock Scaling the codebook size of vqgan to 100,000 with a utilization rate of 99\%.
\newblock \emph{ArXiv}, abs/2406.11837, 2024{\natexlab{a}}.

\bibitem[Zhu et~al.(2024{\natexlab{b}})Zhu, Li, Zhang, Li, Xu, and Bing]{zhu2024stabilize}
Yongxin Zhu, Bocheng Li, Hang Zhang, Xin Li, Linli Xu, and Lidong Bing.
\newblock Stabilize the latent space for image autoregressive modeling: A unified perspective.
\newblock In \emph{The Thirty-eighth Annual Conference on Neural Information Processing Systems}, 2024{\natexlab{b}}.

\bibitem[Zhu et~al.(2024{\natexlab{c}})Zhu, Su, He, Xu, and Yu]{zhu-etal-2024-generative}
Yongxin Zhu, Dan Su, Liqiang He, Linli Xu, and Dong Yu.
\newblock Generative pre-trained speech language model with efficient hierarchical transformer.
\newblock In \emph{Proceedings of the 62nd Annual Meeting of the Association for Computational Linguistics (Volume 1: Long Papers)}, pages 1764--1775, Bangkok, Thailand, 2024{\natexlab{c}}. Association for Computational Linguistics.

\end{thebibliography}
}

\clearpage
\setcounter{page}{1}
\maketitlesupplementary

\section{Appendix}

\subsection{Experimental Configurations}
\label{appendix:config}

Tab. \ref{tab:config} provides the experimental configurations for both image and audio modalities utilized in this study. For the image modality, the input size is specified as $128 \times 128 \times 3$. The batch size for images is set at 256. The model is trained for a total of 50 epochs. Each image is represented with a quantized sequence length of $16 \times 16$, dividing the input data into a grid of tokens. In terms of optimization, the AdamW optimizer is employed with a constant learning rate of $1e-4$, and no warmup epochs are implemented. The commitment coefficient for images is set to $1.0$. The adversarial coefficient for this modality is established at $0.1$, affecting the training dynamics in the context of adversarial methodologies. Regarding data augmentation, a random horizontal flip is applied to the image inputs, enhancing the robustness of the model.

The audio input size is defined as $24,000 \times 1$, reflecting a one-dimensional audio signal sampled at a rate of $24,000$ Hz (1 second). The batch size for audio data is set at $64$. The model undergoes a training duration of $50$ epochs. The optimization settings remain consistent, utilizing the AdamW optimizer and a constant learning rate of $1e-4$ with no warmup epochs. The commitment coefficient for audio is set to $1000.0$ and the adversarial coefficient is set at 1.0, which is the same as WavTokenizer.

\begin{table}[htbp]
\centering
\begin{tabular}	{l|l|l}
\toprule
Config & Image & Audio \\
\midrule
    inputs & pixels & window size \\
    input size & 128 $\times$ 128 $\times$ 3 & 24,000 $\times$ 1 \\
	batch size & 256  & 64  \\
	training epochs & 50 & 50  \\
	quantized sequence length & 16 $\times$ 16  & 75 \\
	\textbf{optimization} \\
	optimizer & AdamW  & AdamW \\ 
	learning rate & 1e-4 & 1e-4 \\
	learning rate schedule & constant & constant \\
	warmup epochs & 0  & 0 \\
    commitment coefficient & 1.0 & 1000.0 \\
    adversarial coefficient & 0.1 & 1.0 \\
	\textbf{data augmentations} \\
    random horizontal flip & true & false \\
	\bottomrule
\end{tabular}
\caption{Experimental configurations on image and audio.}
\label{tab:config}
\end{table}

\subsection{Loss Curve}
\label{appendix:loss}

\begin{figure}[htb]
    \centering
    \includegraphics[width=1.0\columnwidth]{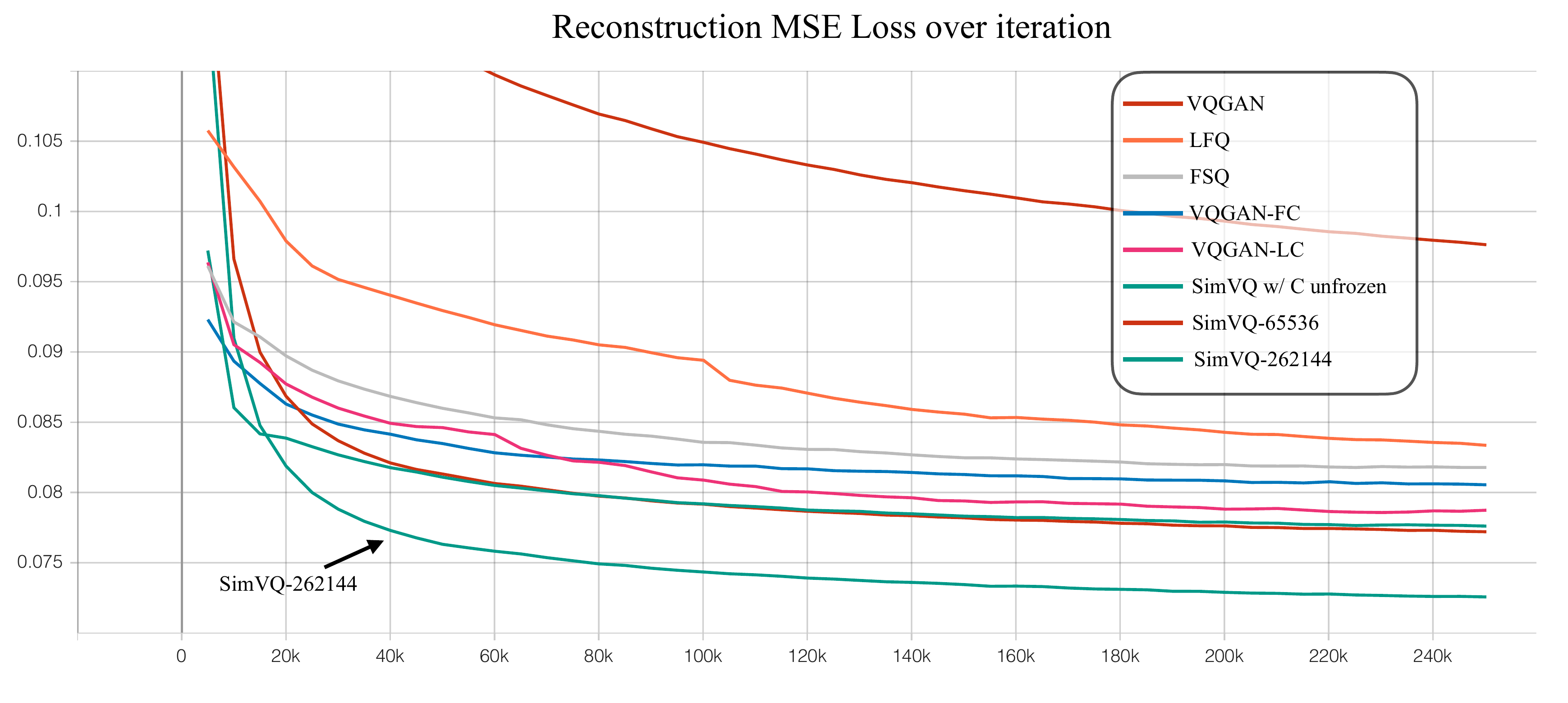}
    \caption{The loss curve over epochs of different models on the validation dataset.}
    \label{fig:loss}
\end{figure}

\subsection{Codebook Distribution}
\label{appendix:freq}

\begin{figure}[htb]
    \centering
    \includegraphics[width=1.0\columnwidth]{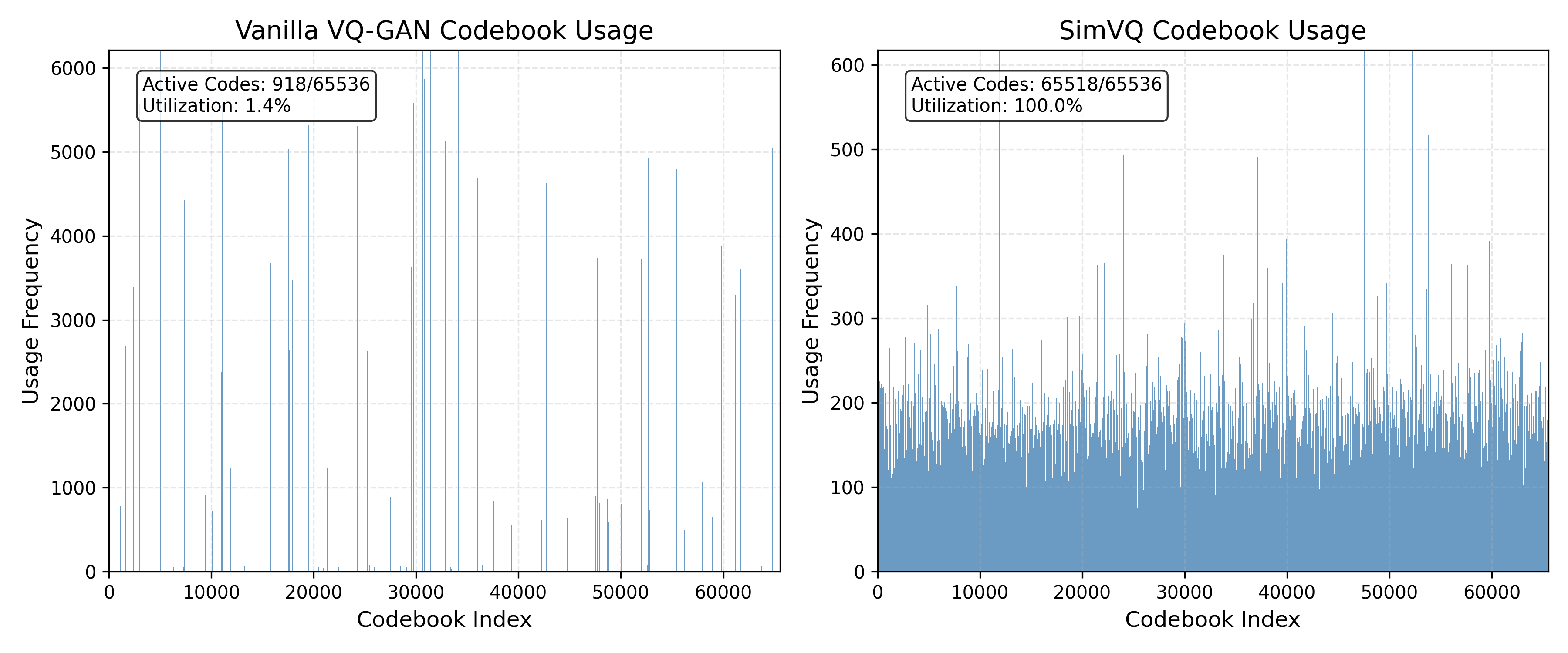}
    \caption{The frequency of codebook on ImageNet validation set.}
    \label{fig:freq}
\end{figure}

\subsection{Qualitative Cases}
\label{appendix:cases}

We provide image and audio cases of SimVQ with various codebook sizes below.

\begin{figure*}[h]
    \centering
    {\small
    \makebox[0.16\textwidth]{Origin}%
    \makebox[0.16\textwidth]{vanilla VQ 65,536}%
    \makebox[0.16\textwidth]{SimVQ 1,024}%
    \makebox[0.16\textwidth]{SimVQ 8,192}%
    \makebox[0.16\textwidth]{SimVQ 65,536}%
    \makebox[0.16\textwidth]{SimVQ 262,144}
    }
    \vspace{5pt} 
    
    \begin{minipage}{0.15\textwidth}
        \centering
        \includegraphics[width=\linewidth]{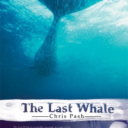}
    \end{minipage}
    \begin{minipage}{0.15\textwidth}
        \centering
        \includegraphics[width=\linewidth]{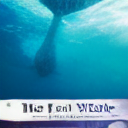}
    \end{minipage}
    \begin{minipage}{0.15\textwidth}
        \centering
        \includegraphics[width=\linewidth]{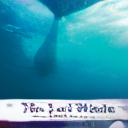}
    \end{minipage}
    \begin{minipage}{0.15\textwidth}
        \centering
        \includegraphics[width=\linewidth]{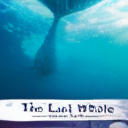}
    \end{minipage}
    \begin{minipage}{0.15\textwidth}
        \centering
        \includegraphics[width=\linewidth]{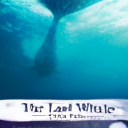}
    \end{minipage}
    \begin{minipage}{0.15\textwidth}
        \centering
        \includegraphics[width=\linewidth]{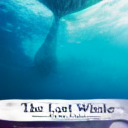}
    \end{minipage}

    \vspace{10pt}
    
    \begin{minipage}{0.15\textwidth}
        \centering
        \includegraphics[width=\linewidth]{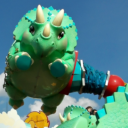}
    \end{minipage}
    \begin{minipage}{0.15\textwidth}
        \centering
        \includegraphics[width=\linewidth]{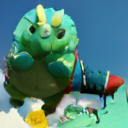}
    \end{minipage}
    \begin{minipage}{0.15\textwidth}
        \centering
        \includegraphics[width=\linewidth]{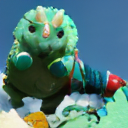}
    \end{minipage}
    \begin{minipage}{0.15\textwidth}
        \centering
        \includegraphics[width=\linewidth]{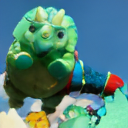}
    \end{minipage}
    \begin{minipage}{0.15\textwidth}
        \centering
        \includegraphics[width=\linewidth]{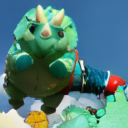}
    \end{minipage}
    \begin{minipage}{0.15\textwidth}
        \centering
        \includegraphics[width=\linewidth]{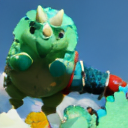}
    \end{minipage}

    \begin{minipage}{0.15\textwidth}
        \centering
        \includegraphics[width=\linewidth]{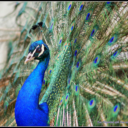}
    \end{minipage}
    \begin{minipage}{0.15\textwidth}
        \centering
        \includegraphics[width=\linewidth]{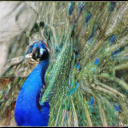}
    \end{minipage}
    \begin{minipage}{0.15\textwidth}
        \centering
        \includegraphics[width=\linewidth]{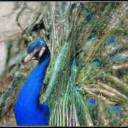}
    \end{minipage}
    \begin{minipage}{0.15\textwidth}
        \centering
        \includegraphics[width=\linewidth]{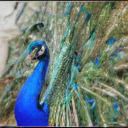}
    \end{minipage}
    \begin{minipage}{0.15\textwidth}
        \centering
        \includegraphics[width=\linewidth]{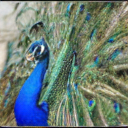}
    \end{minipage}
    \begin{minipage}{0.15\textwidth}
        \centering
        \includegraphics[width=\linewidth]{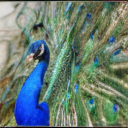}
    \end{minipage}

    \begin{minipage}{0.15\textwidth}
        \centering
        \includegraphics[width=\linewidth]{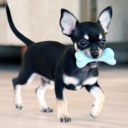}
    \end{minipage}
    \begin{minipage}{0.15\textwidth}
        \centering
        \includegraphics[width=\linewidth]{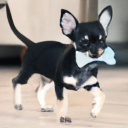}
    \end{minipage}
    \begin{minipage}{0.15\textwidth}
        \centering
        \includegraphics[width=\linewidth]{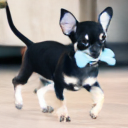}
    \end{minipage}
    \begin{minipage}{0.15\textwidth}
        \centering
        \includegraphics[width=\linewidth]{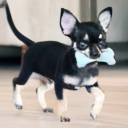}
    \end{minipage}
    \begin{minipage}{0.15\textwidth}
        \centering
        \includegraphics[width=\linewidth]{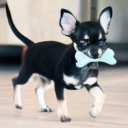}
    \end{minipage}
    \begin{minipage}{0.15\textwidth}
        \centering
        \includegraphics[width=\linewidth]{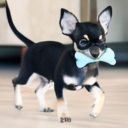}
    \end{minipage}

    \begin{minipage}{0.15\textwidth}
        \centering
        \includegraphics[width=\linewidth]{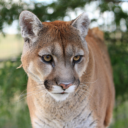}
    \end{minipage}
    \begin{minipage}{0.15\textwidth}
        \centering
        \includegraphics[width=\linewidth]{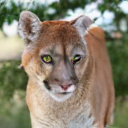}
    \end{minipage}
    \begin{minipage}{0.15\textwidth}
        \centering
        \includegraphics[width=\linewidth]{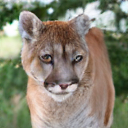}
    \end{minipage}
    \begin{minipage}{0.15\textwidth}
        \centering
        \includegraphics[width=\linewidth]{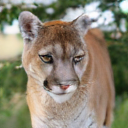}
    \end{minipage}
    \begin{minipage}{0.15\textwidth}
        \centering
        \includegraphics[width=\linewidth]{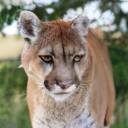}
    \end{minipage}
    \begin{minipage}{0.15\textwidth}
        \centering
        \includegraphics[width=\linewidth]{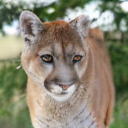}
    \end{minipage}

    \begin{minipage}{0.15\textwidth}
        \centering
        \includegraphics[width=\linewidth]{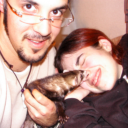}
    \end{minipage}
    \begin{minipage}{0.15\textwidth}
        \centering
        \includegraphics[width=\linewidth]{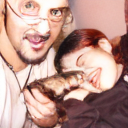}
    \end{minipage}
    \begin{minipage}{0.15\textwidth}
        \centering
        \includegraphics[width=\linewidth]{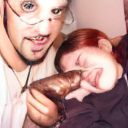}
    \end{minipage}
    \begin{minipage}{0.15\textwidth}
        \centering
        \includegraphics[width=\linewidth]{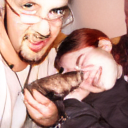}
    \end{minipage}
    \begin{minipage}{0.15\textwidth}
        \centering
        \includegraphics[width=\linewidth]{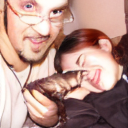}
    \end{minipage}
    \begin{minipage}{0.15\textwidth}
        \centering
        \includegraphics[width=\linewidth]{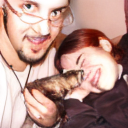}
    \end{minipage}

    \begin{minipage}{0.15\textwidth}
        \centering
        \includegraphics[width=\linewidth]{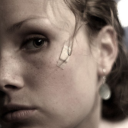}
    \end{minipage}
    \begin{minipage}{0.15\textwidth}
        \centering
        \includegraphics[width=\linewidth]{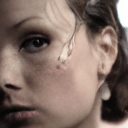}
    \end{minipage}
    \begin{minipage}{0.15\textwidth}
        \centering
        \includegraphics[width=\linewidth]{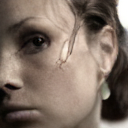}
    \end{minipage}
    \begin{minipage}{0.15\textwidth}
        \centering
        \includegraphics[width=\linewidth]{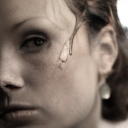}
    \end{minipage}
    \begin{minipage}{0.15\textwidth}
        \centering
        \includegraphics[width=\linewidth]{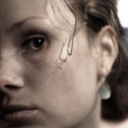}
    \end{minipage}
    \begin{minipage}{0.15\textwidth}
        \centering
        \includegraphics[width=\linewidth]{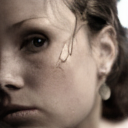}
    \end{minipage}

    \begin{minipage}{0.15\textwidth}
        \centering
        \includegraphics[width=\linewidth]{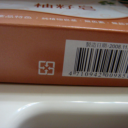}
    \end{minipage}
    \begin{minipage}{0.15\textwidth}
        \centering
        \includegraphics[width=\linewidth]{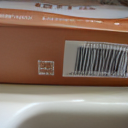}
    \end{minipage}
    \begin{minipage}{0.15\textwidth}
        \centering
        \includegraphics[width=\linewidth]{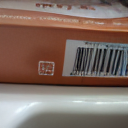}
    \end{minipage}
    \begin{minipage}{0.15\textwidth}
        \centering
        \includegraphics[width=\linewidth]{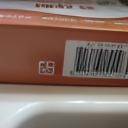}
    \end{minipage}
    \begin{minipage}{0.15\textwidth}
        \centering
        \includegraphics[width=\linewidth]{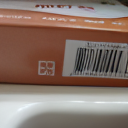}
    \end{minipage}
    \begin{minipage}{0.15\textwidth}
        \centering
        \includegraphics[width=\linewidth]{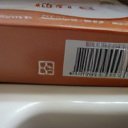}
    \end{minipage}
    
    \caption{Image reconstruction samples with different codebook sizes.}
\end{figure*}

\begin{figure*}[h]
    \centering
    \includegraphics[width=2.0\columnwidth]{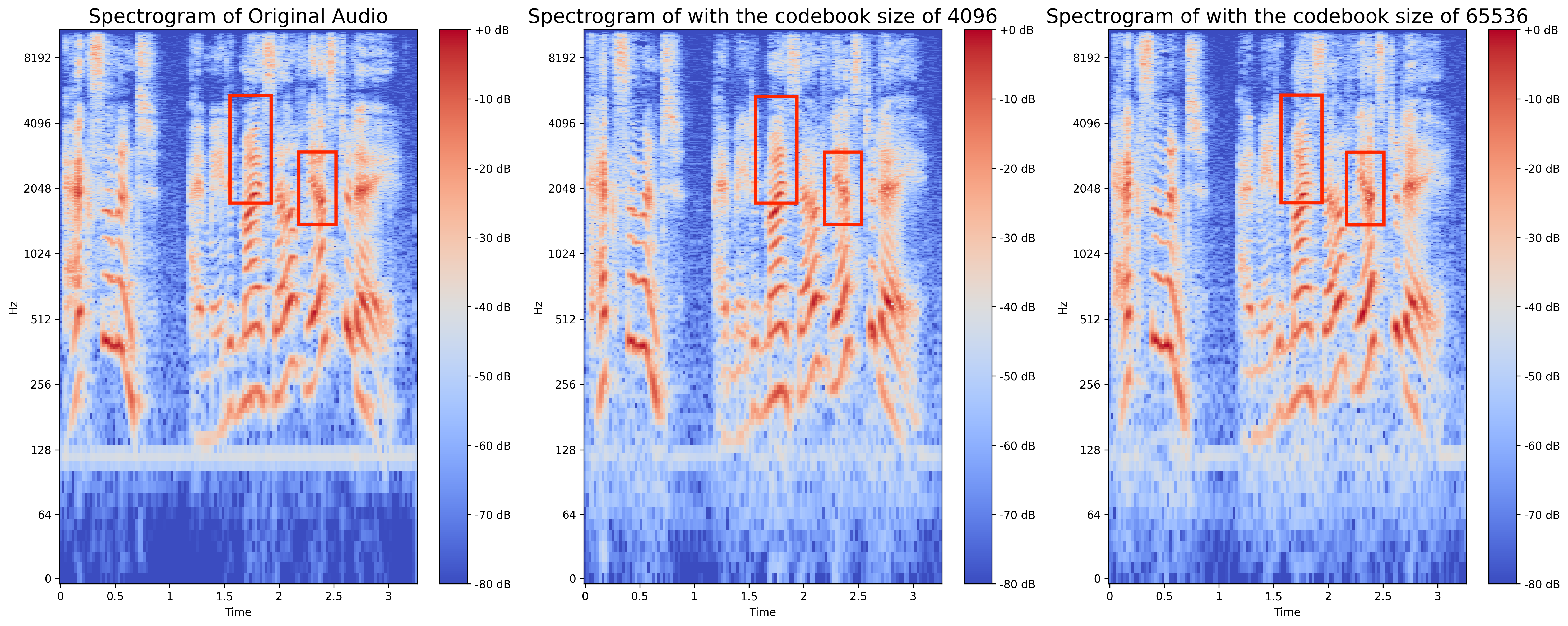}
    \caption{The spectrogram of audio reconstruction samples with different codebook sizes.}
    \label{fig:spec}
\end{figure*}

\begin{figure*}[h]
    \centering
    \includegraphics[width=2.0\columnwidth]{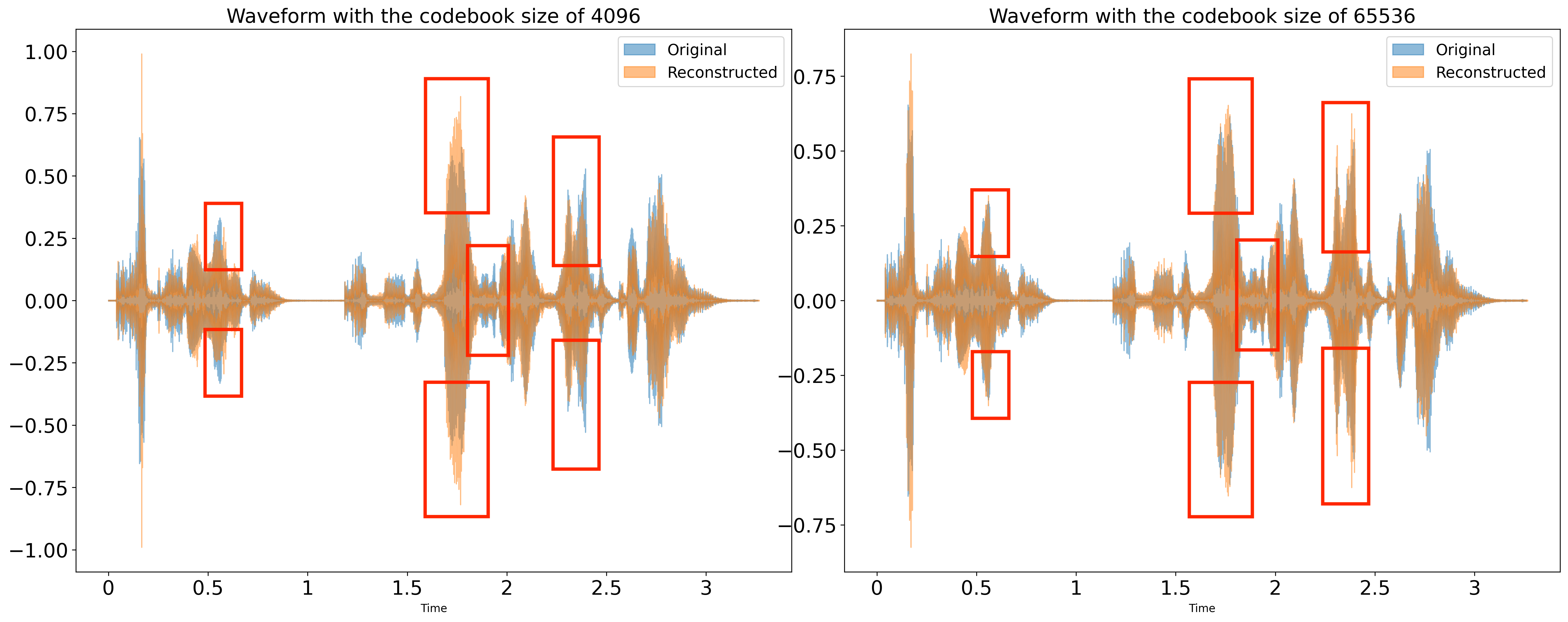}
    \caption{The waveform of audio reconstruction samples with different codebook sizes.}
    \label{fig:waveform}
\end{figure*}

\end{document}